\documentclass[letterpaper, 10 pt, conference]{ieeeconf}  

\IEEEoverridecommandlockouts                              

\usepackage{times}
\usepackage{epsfig}
\usepackage{graphicx}
\usepackage{epstopdf}
\usepackage{subfigure}
\usepackage{multirow}

\usepackage{amsmath}
\usepackage{amssymb}

\usepackage{algorithm}
\usepackage{algorithmicx}
\usepackage{algpseudocode}

\title{\LARGE \bf
Synthetic Data Generation and Adaption for Object Detection in Smart Vending Machines
}

\author{Kai Wang$^{1}$  Fuyuan Shi$^{1}$  Wenqi Wang$^{1}$  Yibing Nan$^{1}$
 Shiguo Lian$^{1}$
\thanks{$^{1}$ All the authors are with CloudMinds Technologies Inc., Beijing 100102, China.
        {\tt\small kai.wang,fuyuan.shi,wenqi.wang,
        charlie.nan,scott.lian@cloudminds.com}}%
}

\begin{document}

\maketitle
\thispagestyle{empty}
\pagestyle{empty}

\begin{abstract}
This paper presents an improved scheme for the generation and adaption of synthetic images for the training of deep Convolutional Neural Networks(CNNs) to perform the object detection task in smart vending machines. While generating synthetic data has proved to be effective for complementing the training data in supervised learning methods, challenges still exist for generating virtual images which are similar to those of the complex real scenes and minimizing redundant training data. To solve these problems, we consider the simulation of cluttered objects placed in a virtual scene and the wide-angle camera with distortions used to capture the whole scene in the data generation process, and post-processed the generated images with a elaborately-designed generative network to make them more similar to the real images. Various experiments have been conducted to prove the efficiency of using the generated virtual images to enhance the detection precision on existing datasets with limited real training data and the generalization ability of applying the trained network to datasets collected in new environment.
\end{abstract}

\section{Introduction}\label{sec:intro}

Smart vending machines are emerging recently due to its convenience and low maintenance cost. Vision-based object detection is usually used in those machines for monitoring the changes of products and is thus one of the most important supporting functions. While deep Convolutional Neural Networks (CNNs) has already shown its advantages in object detection tasks, manual collection and annotation of the large amount of training data are still notably resource and time consuming.
\begin{figure*}[htbp]%
\centering
\includegraphics[scale=0.6]{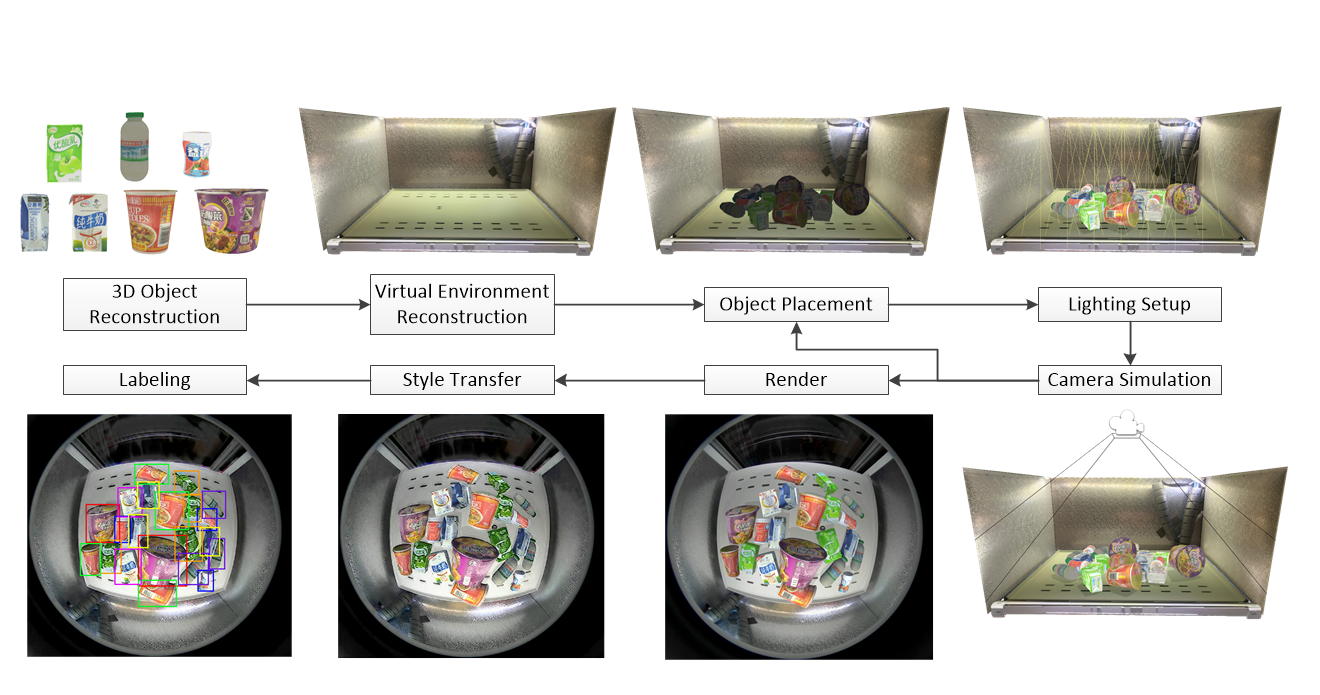}
\caption{The overall workflow of the proposed data generation and adaption method. The 3D objects and interior of the vending machine are first reconstructed. The objects are then arranged in complex layout in the environment. Virtual lights and cameras are next set up and may vary during the generation process. The virtual images are rendered with these settings and their styles are then transferred to better approximate the real data. Bounding boxes for the objects in the generated images are automatically labeled after the image adaption.}
\label{fig:workflow}
\end{figure*}
In recent years, there have been some works trying to solve this problem by generating synthetic training data through virtual environment~\cite{MW16, RB17, TP18}. They render 3D digital models reconstructed or taken from existing dataset using modern computer graphics techniques and create a huge amount of synthetic images by varing the key components in the virtual scene including background texture, lighting, camera parameters, etc. Various training strategies are then adopted to make use of both the synthetic images and the real ones taken in the test environment to enhance the performance of detection. The goal of these approaches is to first make the generated data cover as much sample space as possible and then utilize the real data to compensate for the missing part during training. Although they have proved to be effective on alleviating the insufficient data problem for supervised learning, there still exist some problems:

First, although a variety of factors have been considered to make the rendered images similar to the real image sets in specific scenarios (eg. the interior of a refrigerator), some challenging factors which usually occur in practical systems were not considered. For example, there are only a limited number of objects overlayed onto a background texture, and the generated images may not be realistic when more objects get involved or pose of the camera changes. Meanwhile, a wide-angle camera with distortions is often employed in practice to make sure all the objects in the scene be captured, while few works have addressed this issue for the generation of such images.

Meanwhile, while existing works focus on creating a huge amount of synthetic data and adapt it to the test data set by fusing with the real images during the training process, little efforts have been made to further improve the virtual data itself to make its distribution closer to that of the real data before training. As a result, some synthetic data will be redundant and the performance of the trained network with this data will be limited.

To solve these problem, we proposed an automatic data generation and adaption method for creating and improving the synthetic data used in the training of object detection in smart vending machines. Different from existing works, we considered more realistic scenes with complex object layout and camera models, and improved the virtually rendered images in the data generation process instead of leaving the data adaption task to the training process. Specifically, we first reconstruct the interior of the vending machine as well as the 3D high-quality models of the real products. Then we put the reconstructed models into the virtual scene with complicated layout and various lighting conditions in order to simulate the real case. We also calibrated the wide-angle camera with distortions and simulate it in the virtual scene. Next we rendered the virtual scenes by permuting the parameters around the true values and varying the objects and environment conditions. After getting the rendered images, we further adapt them to the real images through a generative network. To prove the efficiency of our approach, we conducted a series of experiments on various datasets, and showed that the generated and adapted synthetic data can effectively help to improve the detection precision on existing test sets and also generalize well on new datasets.

The contributions of this work include:
\begin{itemize}
  \item An automatic pipeline on effectively producing synthetic training data for detecting and recognizing the number and type of the products in smart vending machines.
  \item An improved scheme for generating synthetic images of realistic environment with complex objects layout and camera models.
  \item A novel approach for adapting the synthetic images to the real datasets to improve the usability of the virtual data.
\end{itemize}

The rest of the paper is organized as follows: Section~\ref{sec:review} reviews the related works on synthetic data generation and object detection. Section~\ref{sec:data} presents the workflow and detailed schemes for generating and adapting the synthetic data. Experimental results are shown and discussed in Section~\ref{sec:exp}. Section~\ref{sec:conc} gives the conclusion.

\section{Related work}\label{sec:review}

\subsection{Data generation}

Synthetic data has been widely used in the training of deep networks for various tasks, including object detection~\cite{LV15,RB17}, pose estimation~\cite{PJ12, HB15}, semantic segmentation~\cite{HP15,RS16}, robotic control~\cite{TF17, MB17}, autonomous driving~\cite{CAR, Air}, etc., and proved to be a useful source for data augmentation. These works can be classified into two categories according to the correspondence between the virtual data generation environment and the real test environment.

For many applications and tasks like robotic control and autonomous driving, it is not quite necessary to require the synthetic training data and test data to have the same distributions. Therefore, the synthetic data can be produced in a virtual environment which is different from the real one. For example, the works~\cite{PS15,LV15} tried to put the CAD models obtained from public datasets onto complex background textures and render them with arbitrary camera and lighting parameters to create the virtual dataset for the training of object detection. Similar works~\cite{MK16,AM17,TP18} created synthetic images containing cars by rendering existing car models onto natural scene images to help enriching the dataset for the detection and segmentation of cars.~\cite{HB15} used digital human models to create synthetic images for pedestrian detection tasks. The CAD models are also used in robotic controls to help training the robot to get the pose of the target to grasp~\cite{TF17,MB17}. There are also some works~\cite{HP15, HK18, RS16, CAR, Air} proposed to create a totally virtual 3D environment, from which various data can be generated for different tasks including object detection, semantic segmentation, disparity estimation, etc. The other reason why the simulated data generated has less overlap with the real data in these works is that they targeted on tasks in more general scenes which may differ a lot among one another.

The other type of works try to recover the real scene's information and then generate the virtual data from the reconstructed virtual scene. For example, there have been some works~\cite{DC17, Mat, AS17} proposed to reconstruct the 3D indoor environment and create various types of training data for different tasks. The test of the trained network can be performed on the data collected in the real environment whose amounts are limited though. The works~\cite{RB17} and~\cite{MW16} used synthetic data for the training of the product recognition tasks in some specific environment like the refrigerator or shelf. The former one used the CAD models which are similar to the real products while the latter used the reconstructed models which have higher precisions. Both of them tried to recover the information of the real environment (such as the camera parameters) and added some perturbations to generate a wider range of synthetic data. However, the layout of their environment is relatively simple and more challenging factors like complicated camera model and object poses were not addressed. Meanwhile, they chose to handle fix the gap between the virtual and real data in the training process, leading to the generation of huge number of redundant data.

\subsection{Object detection}

Generic object detection predicting the object class as well as a rectangle (called bounding box) containing that object. Since the proposal of R-CNN~\cite{GD14}, a lot of CNN-based detectors have been suggested, including two different kinds of approaches. One regards the detection task as a regression or classification problem, make a fixed number of predictions on grid (one stage)~\cite{RD16, LA16}. The other leverages a proposal network to find objects and then uses a second network to fine-tune these proposals and output a final prediction (two stage)~\cite{GD14,GR15,RH17,KC16}. In this work, PVANET~\cite{KC16}, SSD~\cite{LA16}, and YOLOv3~\cite{RD16} architectures are adapted to evaluate the proposed approach.

\section{Synthetic data generation and adaption}\label{sec:data}

The detection of the objects in a complex environment involves the simulation of both the objects and the background. To generate such data, we follow the workflow described in Figure~\ref{fig:workflow}. First, each type of object is reconstructed to recover the geometry and texture information. Next, the 3D models of the objects are placed randomly in a reconstructed virtual scene. After setting the virtual lightings, the whole virtual environment can then be rendered with a virtual camera whose intrinsic parameters are obtained by calibrating the real camera in the machine. Virtual-to-real image transfer is further implemented to make the synthetic images approximate the real ones better, and labeling is finally done automatically. More synthetic images of the training set can be obtained by varying the parameters of the 3D virtual scene(number and types of objects, lighting types and parameters, background textures, etc.).

The details for the key steps in data generation and annotation are described as follows.

\subsection{3D object reconstruction}

A majority of the objects are initially reconstructed using the 3D scanner provided by Shinning~\cite{Shinning}, and the geometry and textures of the object are further refined to enhance high-quality digital models. The scanner adopts the RGB and structural light cameras for the acquisition of the color and depth information respectively, and it is not able to accurately reconstruct objects with specular or transparent materials. To solve this problem, we used the method proposed in~\cite{WH18} to first transform the specular material into a diffuse with a deep neural network to proceed the reconstruction, and restore the specular material afterwards. As there is still a lack of stable method for the automatic reconstruction of transparent objects, we did not include that type of objects during the experiments. Although this could be alleviated by introducing some manual works, it is beyond the scope of this paper and therefore not discussed here. The reconstructed model is represented as a dense triangular mesh. whose bounding box is calculated for the use of arranging the layout of the objects on the virtual holding plane later.

\begin{figure} [htbp]
  \centering
  \subfigure[]{
    \centering
    \label{fig:deform:a} 
    \begin{minipage}[b]{0.13\textwidth}
      \centering
      \includegraphics[scale=0.15]{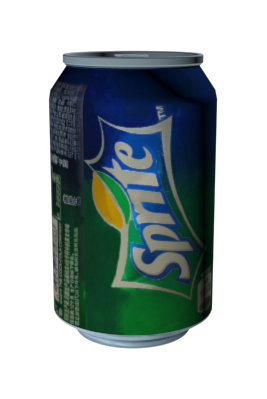}
    \end{minipage}}
  \subfigure[]{
    \centering
    \label{fig:deform:b}
    \begin{minipage}[b]{0.13\textwidth}
      \centering
      \includegraphics[scale=0.12]{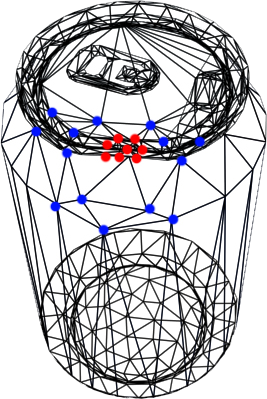}
    \end{minipage}}
  \subfigure[]{
    \centering
    \label{fig:deform:c}
    \begin{minipage}[b]{0.13\textwidth}
      \centering
      \includegraphics[scale=0.15]{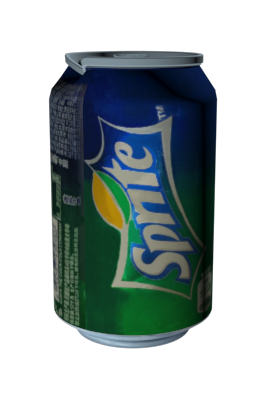}
    \end{minipage}}
  \subfigure[]{
    \centering
    \label{fig:deform:d}
    \begin{minipage}[b]{0.13\textwidth}
      \centering
      \includegraphics[scale=0.15]{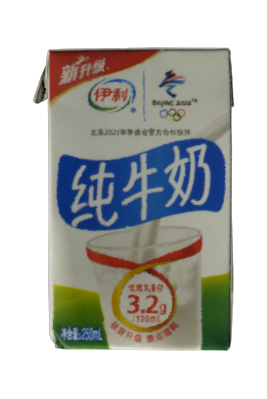}
    \end{minipage}}
  \subfigure[]{
    \centering
    \label{fig:deform:e}
    \begin{minipage}[b]{0.13\textwidth}
      \centering
      \includegraphics[scale=0.12]{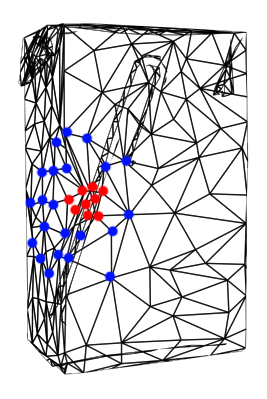}
    \end{minipage}}
  \subfigure[]{
    \centering
    \label{fig:deform:f}
    \begin{minipage}[b]{0.13\textwidth}
      \centering
      \includegraphics[scale=0.15]{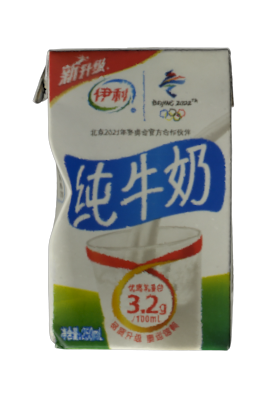}
    \end{minipage}}
  \caption{Examples of model deformation. (a) and (d): The reconstructed models. (b) and (e): The handle (red) and neighbor (blue) vertices of the meshese. (c) and (f): The corresponding deformed models.}
  \label{fig:deform}
\end{figure}

In practice, most of the products are subject to different extents of deformation, due to external forces. As a result, their appearance may probably be different from the scanned digital models. Therefore, we simulate the deformation of the digital model by adding random to its surface. To do that, we employed the surface-based mesh deformation algorithm proposed in~\cite{WZ12}. We first randomly select a group of handle vertices as the centers of the deformed areas (the red vertices in Figure~\ref{fig:deform:b} and Figure~\ref{fig:deform:e}), and their neighbors (the blue vertices in Figure~\ref{fig:deform:b} and Figure~\ref{fig:deform:e}) will deform with the handle vertices. The other vertices of the mesh will keep fixed during the deformation. Specifically, the deformed vertex positions of the mesh are calculated by minimizing the following energy function:

\begin{eqnarray}
\begin{split}
\label{eq:edgelocflexenergy}
& E(V^\prime) = \sum\limits_{i=1}^n{\lambda_i(\sum\limits_{j\in N(i)}{||\tilde{e_{ij}}^\prime -R_i \tilde{e_{ij}}||^2})}\\
& + \sum\limits_{i=1}^n{(1-\lambda_i)||L(\tilde{v_i}^\prime)-T_iL(\tilde{v_i}||^2}\\
& + \sum\limits_{i=m}^n{||\tilde{v_i}^\prime-\tilde{c_i}||^2},
\end{split}
\end{eqnarray}

in which the first and second items represent the first and second-order properties (related to the rigidity and smoothness) of the mesh surface respectively. $\tilde{e_{ij}}^\prime=\tilde{v_i}^\prime-\tilde{v_j}^\prime$ and $\tilde{e_{ij}}=\tilde{v_i}-\tilde{v_j}$ represent the deformed and original edge vectors between vertex $v_i$ and ${v_j}$ respectively. $L(\tilde{v_i}^\prime)$ and  $L(\tilde{v_i}$ represent the Laplacian vectors at vertex $i$ whose computations can be found in~\cite{WZ12}. $R_i$ and $T_i$ are the rotation and transformation matrices which can be solved in an iterative manner. The last items represent the positional constraints for the handle vertices during the deformation. In this equation, $\lambda_i$ is the key variable used to control the rigidity of the physical material of the represented object. A larger value of $\lambda_i$ will be specified if the material is relatively rigid, while a smaller one will be given if it is soft.

For the same types of objects, a same $\lambda_i$ value will be given; while for different objects of the same type, different number of deformation areas will be selected by specifying different groups of handle and fixed vertices. Some examples of the deformed objects can be found in Figure~\ref{fig:deform}. As the system to solve is linear, this algorithm can be finished in real time and it is time-consuming to simulate the deformation of the real object's surface.

\subsection{Virtual environment simulation}

After the objects have been reconstructed, they will be placed into a virtual environment which simulates that interior of a real vending machine. Here we used the BundleFusion~\cite{DN17} to reconstruct the interior structure and textures of the vending machine and further refined it with the method proposed in~\cite{HD17}. The virtual scene together with the reconstructed objects will then be rendered with a virtual camera to generate the synthetic images. We use the Unity3D~\cite{Unity} engine to build up the whole virtual environment and render the images. There are some key factors involved in this process: the layout of the environment and objects, the camera parameters, and the lightings.

Similar to the real environment, the layout of the virtual one generally has a cuboid shape consisting of a flat holding plane for placing the objects. We place the camera on the ceiling of the reconstructed machine and make it face the objects placed on the holding plane.

An important task here is to place the objects on the holding plane in an efficient manner. For the ease of implementing the detection and classification, we allow two general poses of the objects: the standing pose and the lying pose. For the former pose, all objects are made to stand on the holding plane with tops facing the camera, and for the latter one, they are placed to lie on the plane with side views facing the camera. To place the objects with either standing pose or lying pose onto the holding plane, we simplify the problem into placing the excircles of the top or side views of the objects into a rectangle area. The excircles can be easily computed from the bounding boxes of the models.

\begin{figure} [htbp]
  \centering
  \subfigure[]{
    \centering
    \label{fig:layout:a} 
    \begin{minipage}[b]{0.11\textwidth}
      \centering
      \includegraphics[scale=0.1]{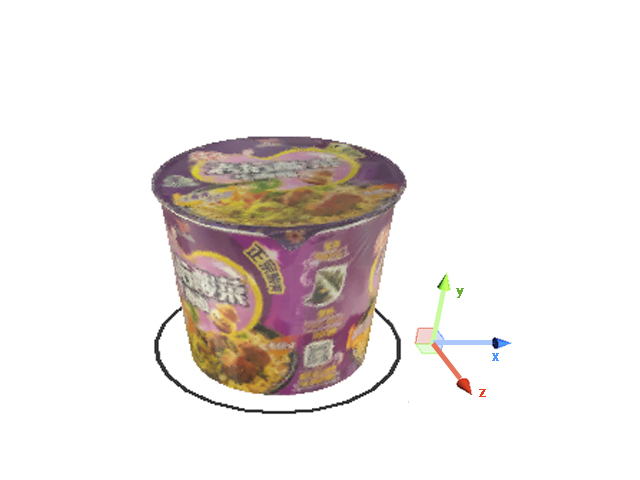}
    \end{minipage}}
  \subfigure[]{
    \centering
    \label{fig:layout:b}
    \begin{minipage}[b]{0.11\textwidth}
      \centering
      \includegraphics[scale=0.1]{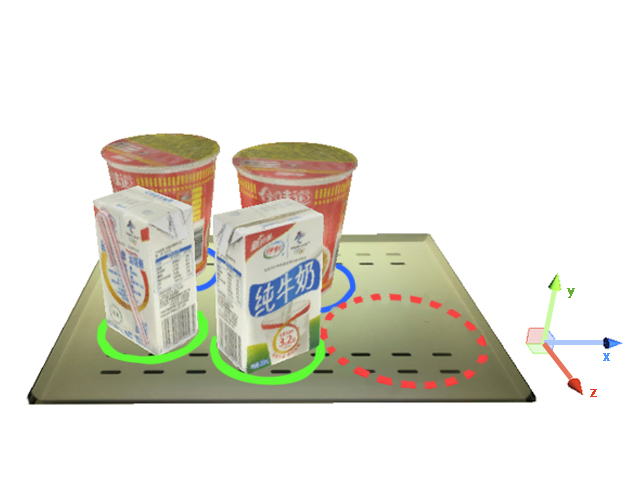}
    \end{minipage}}
  \subfigure[]{
    \centering
    \label{fig:layout:c}
    \begin{minipage}[b]{0.11\textwidth}
      \centering
      \includegraphics[scale=0.1]{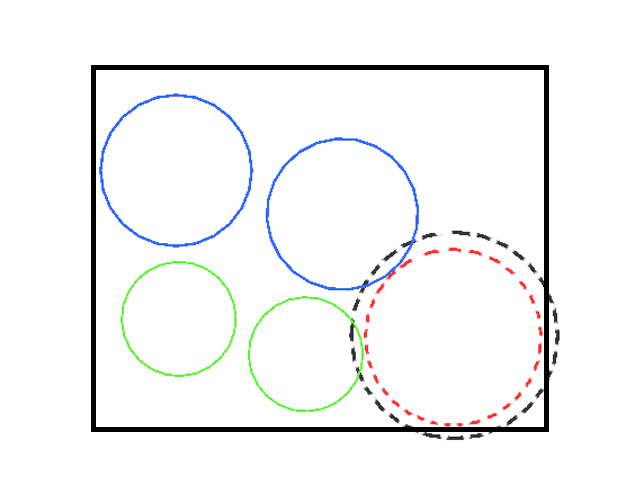}
    \end{minipage}}
  \subfigure[]{
    \centering
    \label{fig:layout:d}
    \begin{minipage}[b]{0.11\textwidth}
      \centering
      \includegraphics[scale=0.1]{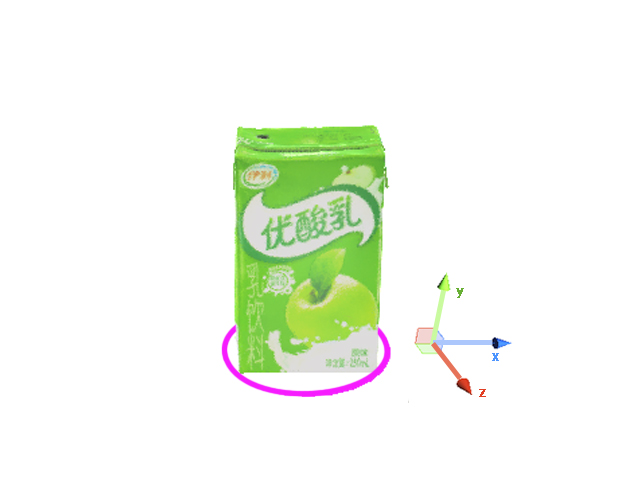}
    \end{minipage}}
  \subfigure[]{
    \centering
    \label{fig:layout:e}
    \begin{minipage}[b]{0.11\textwidth}
      \centering
      \includegraphics[scale=0.1]{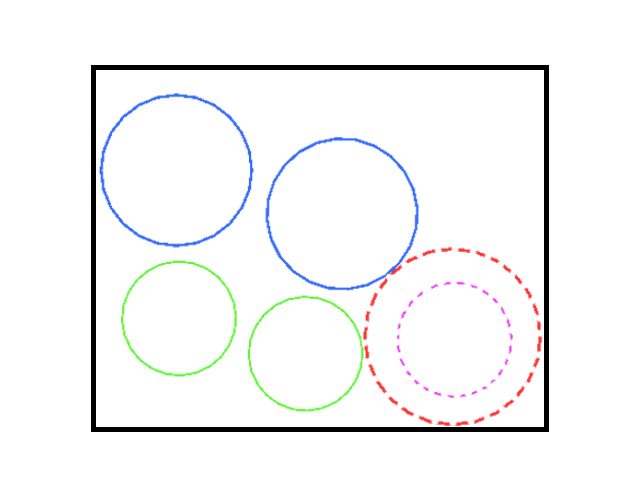}
    \end{minipage}}
  \subfigure[]{
    \centering
    \label{fig:layout:f}
    \begin{minipage}[b]{0.11\textwidth}
      \centering
      \includegraphics[scale=0.1]{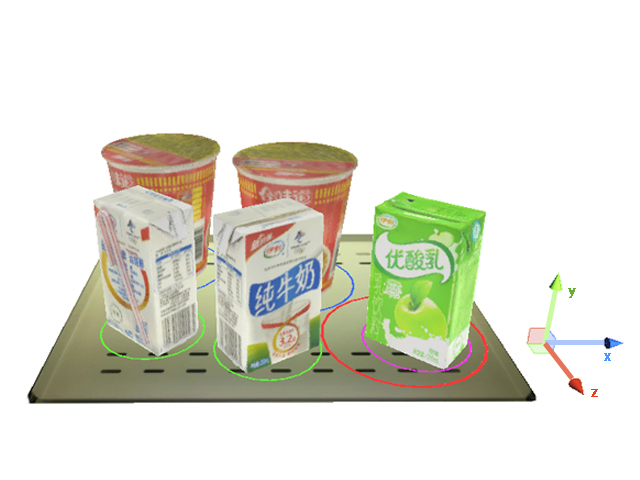}
    \end{minipage}}
  \subfigure[]{
    \centering
    \label{fig:layout:g}
    \begin{minipage}[b]{0.11\textwidth}
      \centering
      \includegraphics[scale=0.1]{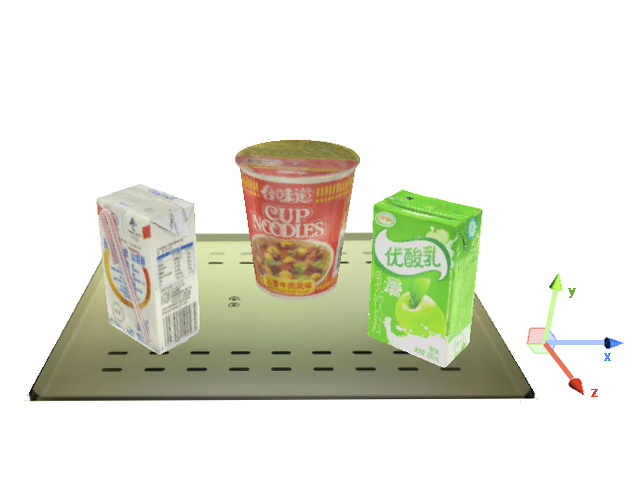}
    \end{minipage}}
  \caption{Examples of placing the virtual objects onto the holding plane. (a) A random object is selected and its excircle (in black) in $Y$ direction is calculated. (b) The incircle (in red) of the remaining part of the holding plane is calculated. (c) The black excircle is bigger than the red incircle, so the selected object cannot be placed onto the holding plane. (d) Another object is randomly selected and its excircle (in pink) is calculated. (e) The pink excircle is smaller than the red circle. (f) The object is rotated and placed into the incircle on the holding plane. (e) A random set of objects is selected from all placed objects.}
  \label{fig:layout}
\end{figure}

To effectively place the objects with abundant types and numbers and meanwhile maintain the randomness of the sample distribution, we follow the algorithm presented in~\ref{alg:place}, in which a random object set $\{b\}$ will be selected from the reconstructed object set $\{A\}$ and placed on the holding plane $P$. Note that in Line 4, we compute the incircle of the remaining part of the plane (represented as a polygon) by constructing the Voronoi Diagram of its edges using the algorithm in~\cite{FO86} and finding the node with maximum distance to the polygon edges. The node position is the desired incircle center and the distance is the radius. Examples of placing onto the holding plane are shown in Figure~\ref{fig:layout}.

\begin{algorithm}[htbp]
\caption{PlaceObjectsonPlane$(\{A\},P,\{b\})$}
\label{alg:place}
\begin{algorithmic}[1]
\While {$\{A\} \ne \emptyset$}
\State select a random object $A_i  \in \{A\}$
\State compute the excircle $E_{A_i}$ of $A_i$ in $Y$ direction
\State compute the incircle $I_P$ of the remaining part of $P$
\If {$Area(E_{A_i})>Area(I_P)$}
\State $\{A\}=\{A\}-A_i$
\Else
\State rotate $A_i$ around $Y$ axis with a random angle $\theta$
\State place $A_i$ into $I_P$
\State Selected object set $\{B\}=\{B\}+A_i$
\EndIf
\EndWhile
\State select $\{b\} \subseteq \ \{B\}$\\
\Return $\{b\}$
\end{algorithmic}
\end{algorithm}

To simulate environment with different lightings, we randomly select a number of point lights (no more than 5) with random intensities and directions and put them inside the cuboid. We also randomly placed a few point lights (no more than 3) outside the cuboid to simulate the lightings coming from the exterior of the vending machine.

\subsection{Virtual camera setup}

After the virtual environment has been set up, we need to render it using a virtual camera. The camera needs to be basically consistent with that of the real fisheye camera. To do that, we first calibrate the real camera using the method proposed in~\cite{KB06} and get the focal length $(f_x, f_y)$, principal points $(c_x, c_y)$ and the fisheye distortion parameters ${k_1, k_2, k_3, k_4}$. As the all 3D engines in modern computer graphics only support the pinhole camera model, we first set the projection matrix $M_{Proj}$ of the virtual camera as:

\begin{eqnarray}\label{eq:proj}
 \left( \begin{array}{cccc} \frac{2f_x}{w} & 0 & \frac{-2(c_x-w/2)}{w} & 0 \\ 0 & \frac{2f_y}{h} & \frac{-2(c_y-h/2)}{h} & 0 \\ 0 & 0 & \frac{-(near+far)}{(far-near)} & \frac{-2*near*far}{(far-near)} \\ 0 & 0 & -1 & 0 \end{array} \right),
\end{eqnarray}

in which $width$ and $height$ represent the width and height of the rendered image in pixels respectively. $near$ and $far$ represent the values of the near and far clipping planes.

After obtaining the projected image, we then unproject each pixel $(p_x, p_y)$ into the camera space through multiplying the inverse of the intrinsic matrix:

\begin{eqnarray}\label{eq:unproj}
\left[ \begin{array}{l}
p_x^c\\
p_y^c\\
p_z^c
\end{array} \right] &=&
\left[ {\begin{array}{*{20}{c}}
{{f_x}}&{{0}}&{{c_x}}\\
{{0}}&{{f_y}}&{{c_y}}\\
{{0}}&{{0}}&{{1}}
\end{array}} \right]^{-1}
\left[ \begin{array}{l}
p_x\\
p_y\\
1
\end{array} \right],
\end{eqnarray}

where ${p_x^c, p_y^c, p_z^c}$ are the unprojected camera coordinates.

The distorted positions $(p_u, p_v)$ can then be calculated through the following equations.

\begin{eqnarray}\label{eq:distort}
  a &=& p_x^c/p_z^c,\nonumber\\
  b &=& p_y^c/p_z^c,\nonumber\\
  \theta &=& atan(a^2+b^2),\nonumber\\
  \theta_d &=& \theta(1+k_1\theta^2+k_2\theta^4+k_3\theta^6+k_4\theta^8),\nonumber\\
  x^\prime &=&   a* \theta_d/(a^2+b^2),\nonumber\\
  y^\prime &=&   b* \theta_d/(a^2+b^2),\nonumber\\
  p_u &=& f_x(x^\prime+ay^\prime)+c_x,\nonumber\\
  p_v &=& f_yy^\prime+c_y.
\end{eqnarray}

Furthermore, to simulate fisheye cameras with different intrinsic parameters, we vary the calibrated intrinsic parameters as well as the camera positions. Before performing the rendering, we also randomly select the background image from a batch of textures on which the 3D objects will be rendered. This way, more distinct rendered images with the same object layout can be obtained and the sample space can thus be enriched.


\subsection{Virtual-to-real image style transfer}\label{sec:data:transfer}

\begin{figure}[htbp]%
\centering
\includegraphics[scale=0.5]{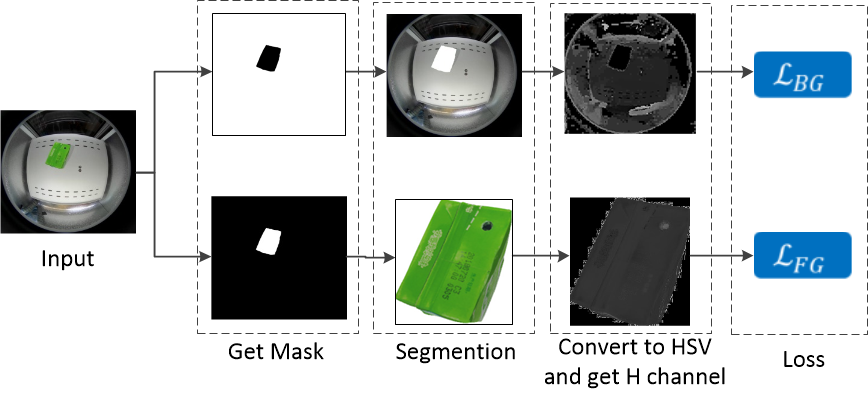}
\caption{The virtual-to-real image style transfer architecture.}
\label{fig:styletransfer}
\end{figure}

Although various factors have been considered during the data generation, there is still distinction between the synthetic image and real image, which may reduce performance. To bridge the domain gap between synthetic image and real image, transfer of image style is made in this step.

Compared with foreground objects, background changes less with parameters of the simulator and real capture scenario. If we perform the style transfer on entire image directly by the Cycle-GAN~\cite{ZP17}, style of background may change too much to be used. For the foreground objects, there are much more difference, such as material, edge, and contrast. Therefore, different from traditional image-to-image transfer, using different parameters to transfer background and foreground is more appropriate.

The original loss function of Cycle-GAN is defined as:

\begin{equation}\label{eq:cyclegan}
\begin{aligned}
L_{style} &= L_{GAN}(G, D_Y, X, Y)\\
& +L_{GAN}(F, D_X, X, Y)\\
& +\lambda_1 L_{cyc}(G,F),
\end{aligned}
\end{equation}

where $X$ and $Y$ represent two datasets. $G$ represents the style mapping function from $X$ to $Y$ and $F$ represents the style mapping function from $Y$ to $X$. $DX$ and $DY$ are the style discriminators for $X$ and $Y$. $L_GAN$ represents the standard adversarial loss, and $L_cyc$ represents the cycle consistency loss. For more details of those loss functions, please refer to the Cycle-GAN~\cite{ZP17}. To transfer foreground objects suitable and reserve the relative invariance of background at same time, we change the loss function to:

\begin{equation}\label{eq:loss}
L_{OD} = L_{style}+\lambda_2 L_{BG}+\lambda_3 L_{FG},
\end{equation}

where $L_{BG}$ denotes the background loss which is similar to original loss $L_{style}$, and $L_{FG}$ denotes the foreground objects loss, and $\lambda_2$ and $\lambda_3$ are the parameters for the trade-off between three losses.

In our scenario, color and profile are critical information for product recognition. These information need to be reserved after transfering as much as possible. In this work, images are converted from RGB color space to HSV color space before computing $L_{FG}$, only the H channel has constraint condition of forward cycle consistency. So, the foreground loss is formulated as follows:

\begin{equation}\label{eq:fg}
\begin{aligned}
& L_{FG} = \mathbb{E}_{x\sim p_{data}(x)}{\left[{\left\|{(G(x)_H-x_H)}\odot{M_{FG}{(x)}}\right\|}_2\right]}\\
& +\mathbb{E}_{y\sim p_{data}(y)}{\left[{\left\|{(F(y)_H-y_H)}\odot{M_{FG}{(y)}}\right\|}_2\right]},
\end{aligned}
\end{equation}

 where $x\sim p_{data}(x)$ and $y\sim p_{data}(y)$ represent data distribution of A and B. And $M_{FG}{(x)}$ denotes the foreground mask of all objects for image $x$. If replace $M_{FG}{(x)}$ to background $M_{BG}{(x)}$ and use forward cycle consistency to RGB images, the $L_{FG}$ definition will be changed to $L_{BG}$:

\begin{equation}\label{eq:bg}
\begin{aligned}
& L_{BG} = \mathbb{E}_{x\sim p_{data}(x)}{\left[{\left\|{((G(x))-x)}\odot{M_{BG}{(x)}}\right\|}_2\right]}\\
& +\mathbb{E}_{y\sim p_{data}(y)}{\left[{\left\|{(F(y)-y)}\odot{M_{BG}{(y)}}\right\|}_2\right]}.
\end{aligned}
\end{equation}

The transfer architecture is shown in Figure~\ref{fig:styletransfer}. For each image sample, background and foreground are segmented. Before the transfer, a background dataset gathered from real scenario is established first. If the sample is a real image, segmentation is performed by calculating difference between real image including objects and corresponding clean template of background image. With regard to synthetic image, segmentation is performed by generation engine directly. Then, $L_{BG}$ and $L_{FG}$ can be computed respectively.

\begin{figure} [htbp]
  \centering
  \subfigure[]{
    \centering
    \label{fig:transres:a} 
    \begin{minipage}[b]{0.14\textwidth}
      \centering
      \includegraphics[scale=0.08]{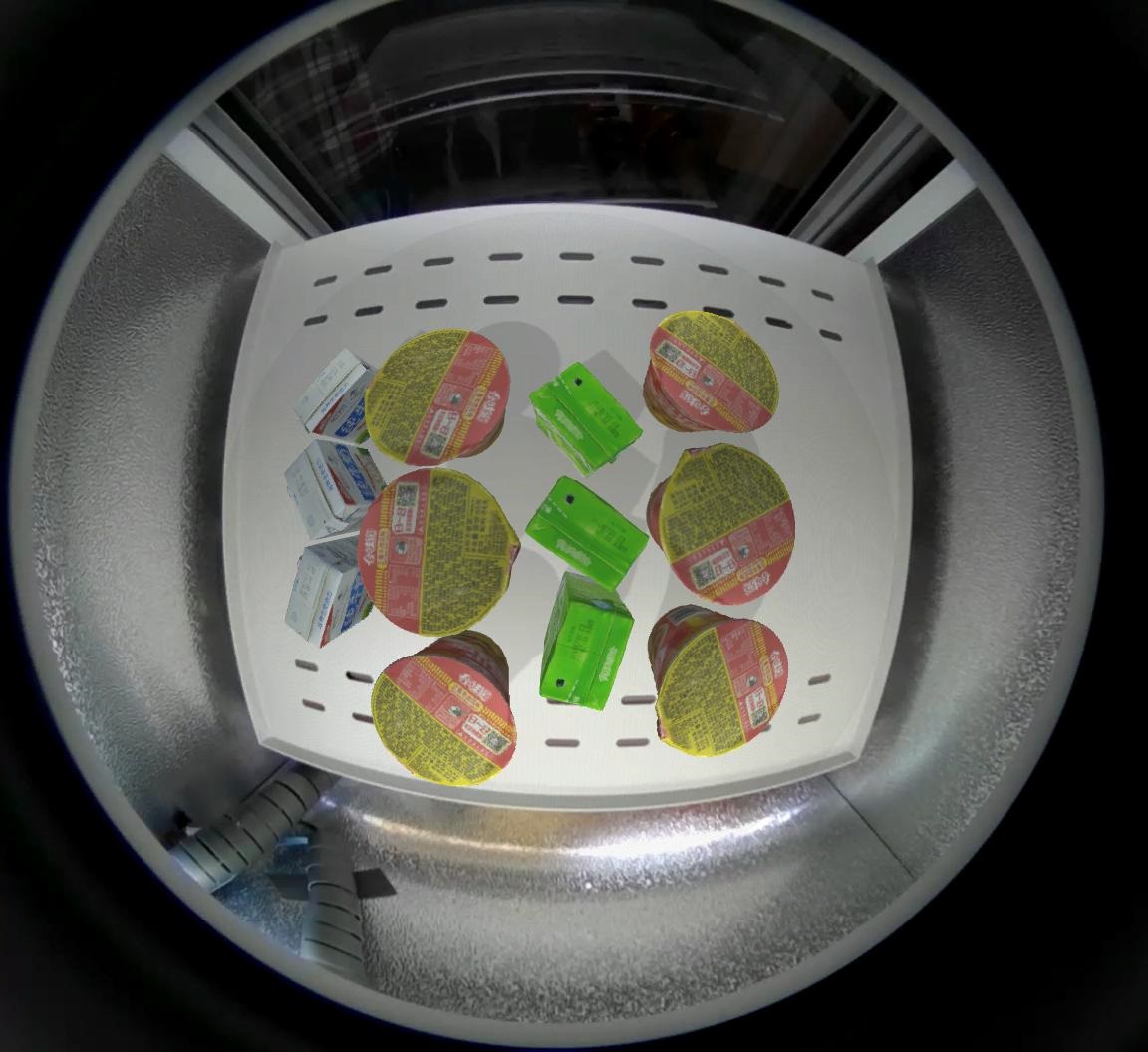}
    \end{minipage}}
  \subfigure[]{
    \centering
    \label{fig:transres:b}
    \begin{minipage}[b]{0.14\textwidth}
      \centering
      \includegraphics[scale=0.08]{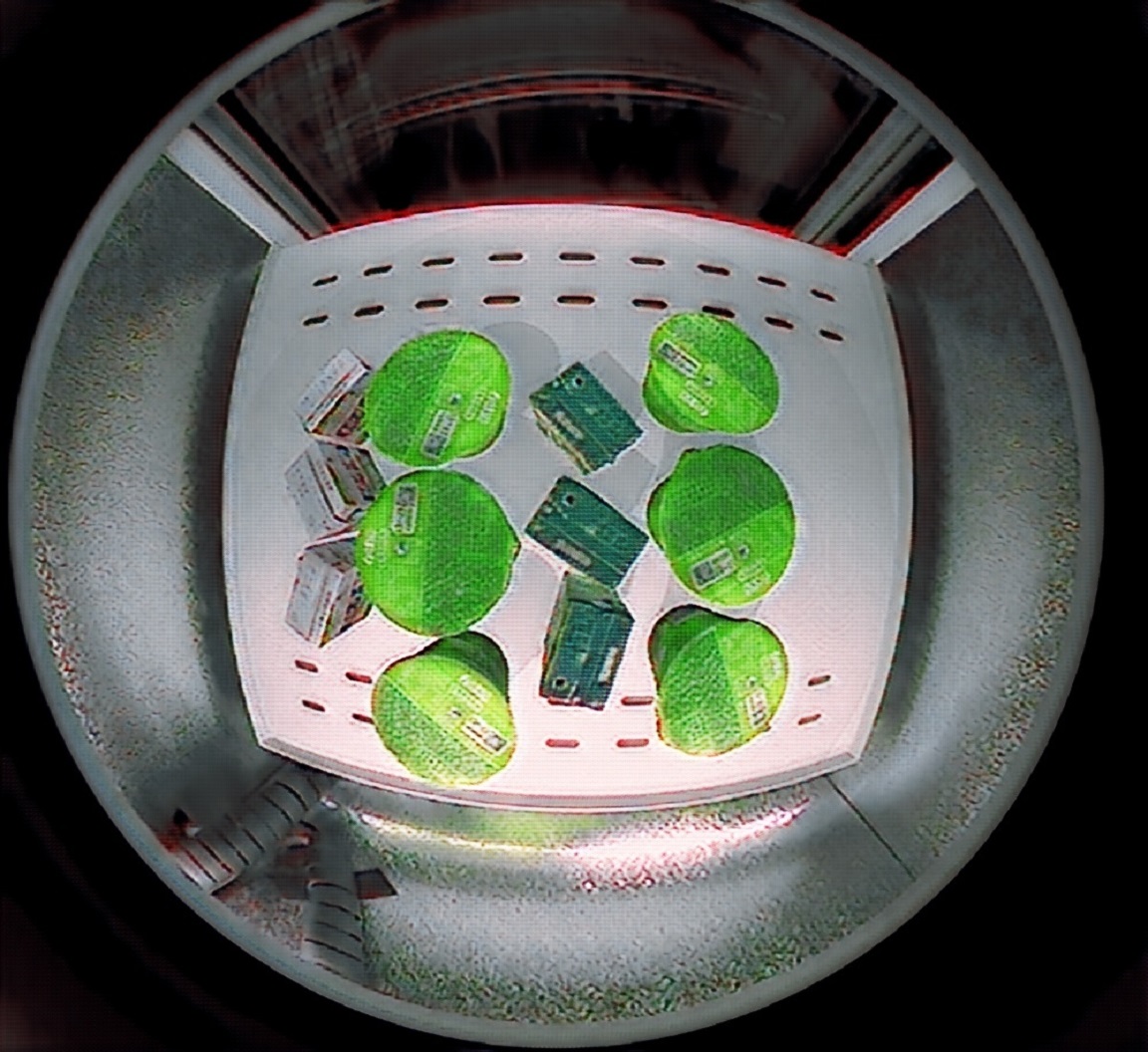}
    \end{minipage}}
  \subfigure[]{
    \centering
    \label{fig:transres:c}
    \begin{minipage}[b]{0.14\textwidth}
      \centering
      \includegraphics[scale=0.08]{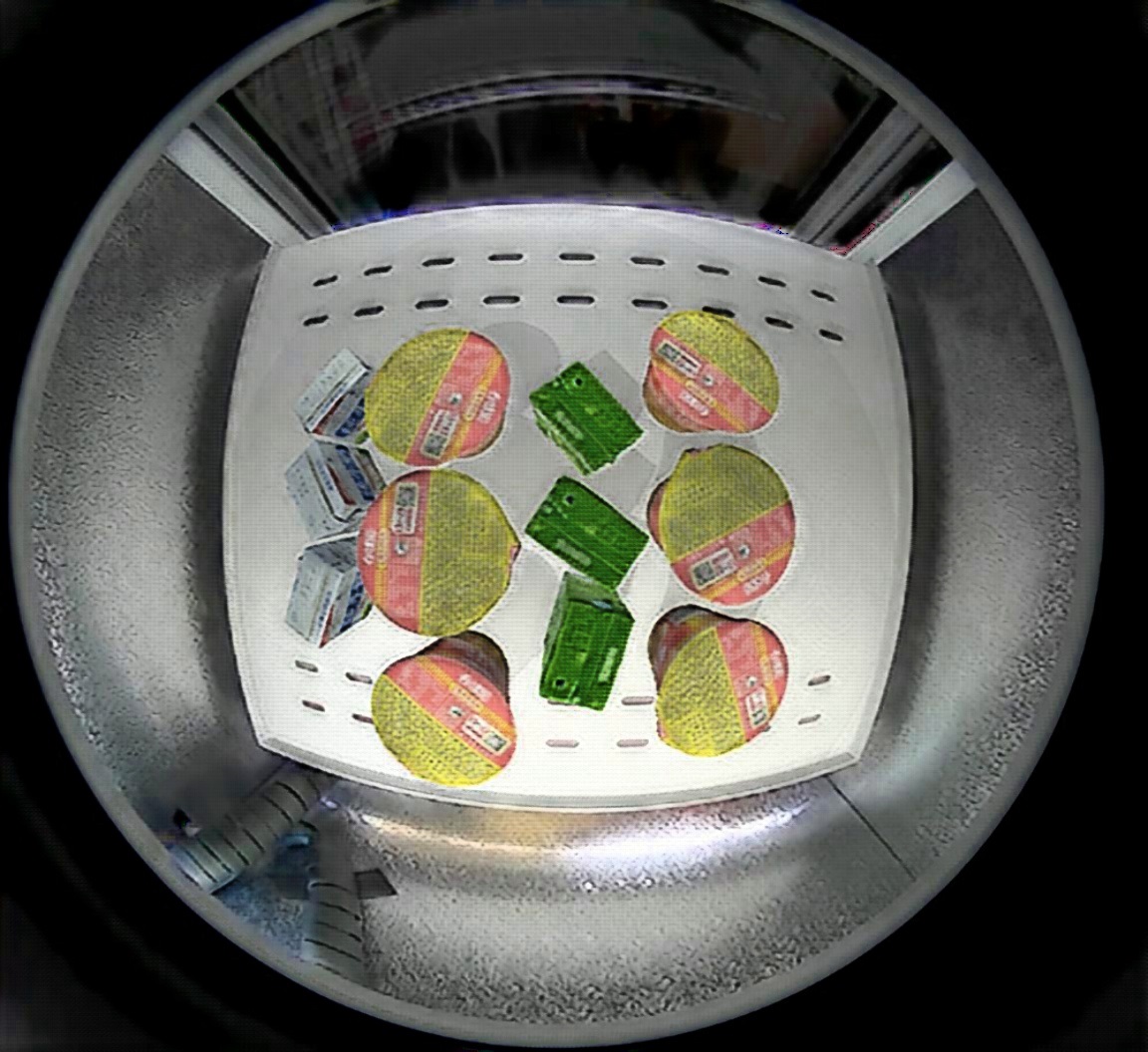}
    \end{minipage}}
  \subfigure[]{
    \centering
    \label{fig:transres:d}
    \begin{minipage}[b]{0.14\textwidth}
      \centering
      \includegraphics[scale=0.08]{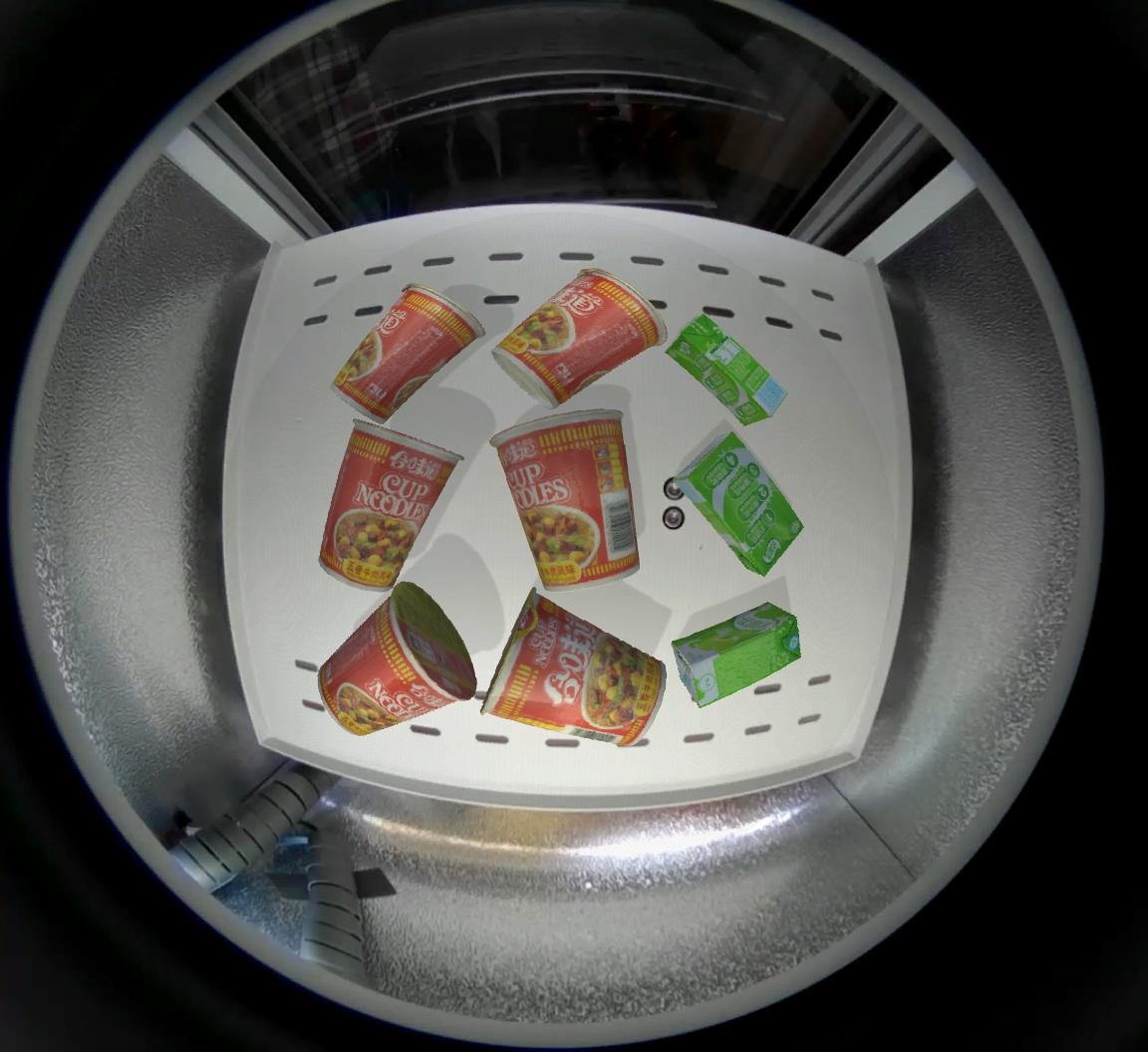}
    \end{minipage}}
  \subfigure[]{
    \centering
    \label{fig:transres:e}
    \begin{minipage}[b]{0.14\textwidth}
      \centering
      \includegraphics[scale=0.08]{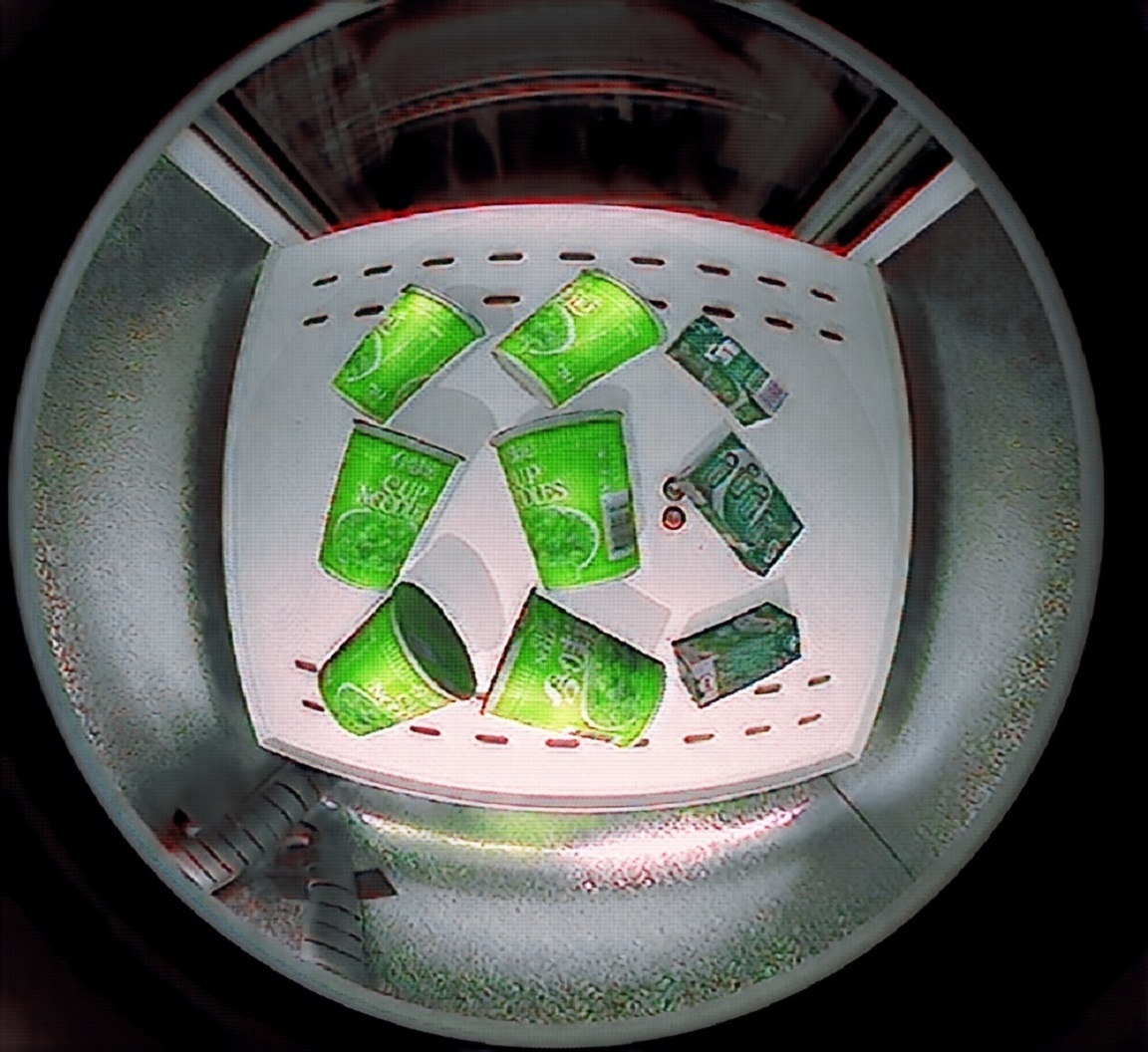}
    \end{minipage}}
  \subfigure[]{
    \centering
    \label{fig:transres:f}
    \begin{minipage}[b]{0.14\textwidth}
      \centering
      \includegraphics[scale=0.08]{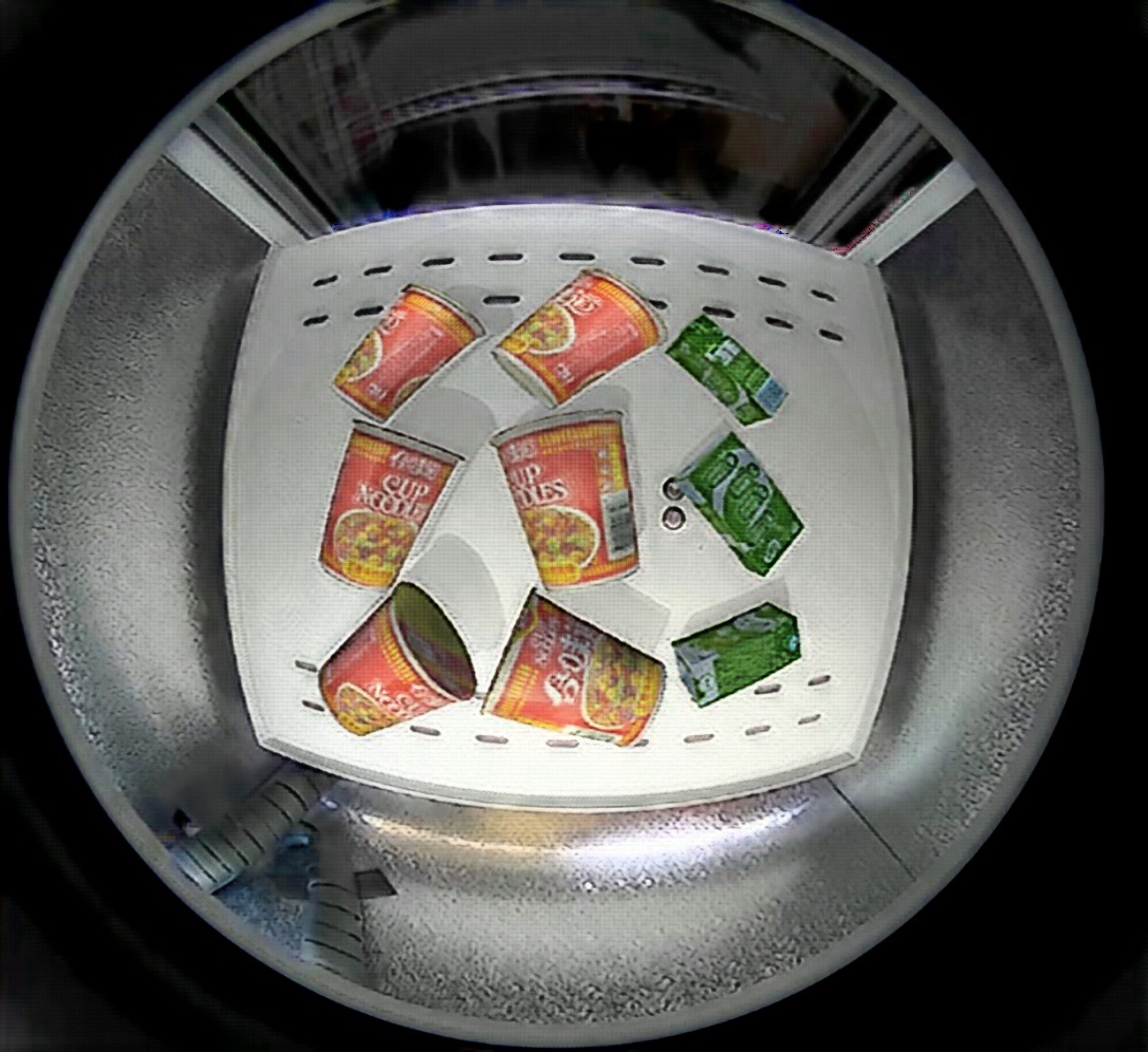}
    \end{minipage}}
  \subfigure[]{
    \centering
    \label{fig:transres:g}
    \begin{minipage}[b]{0.14\textwidth}
      \centering
      \includegraphics[scale=0.08]{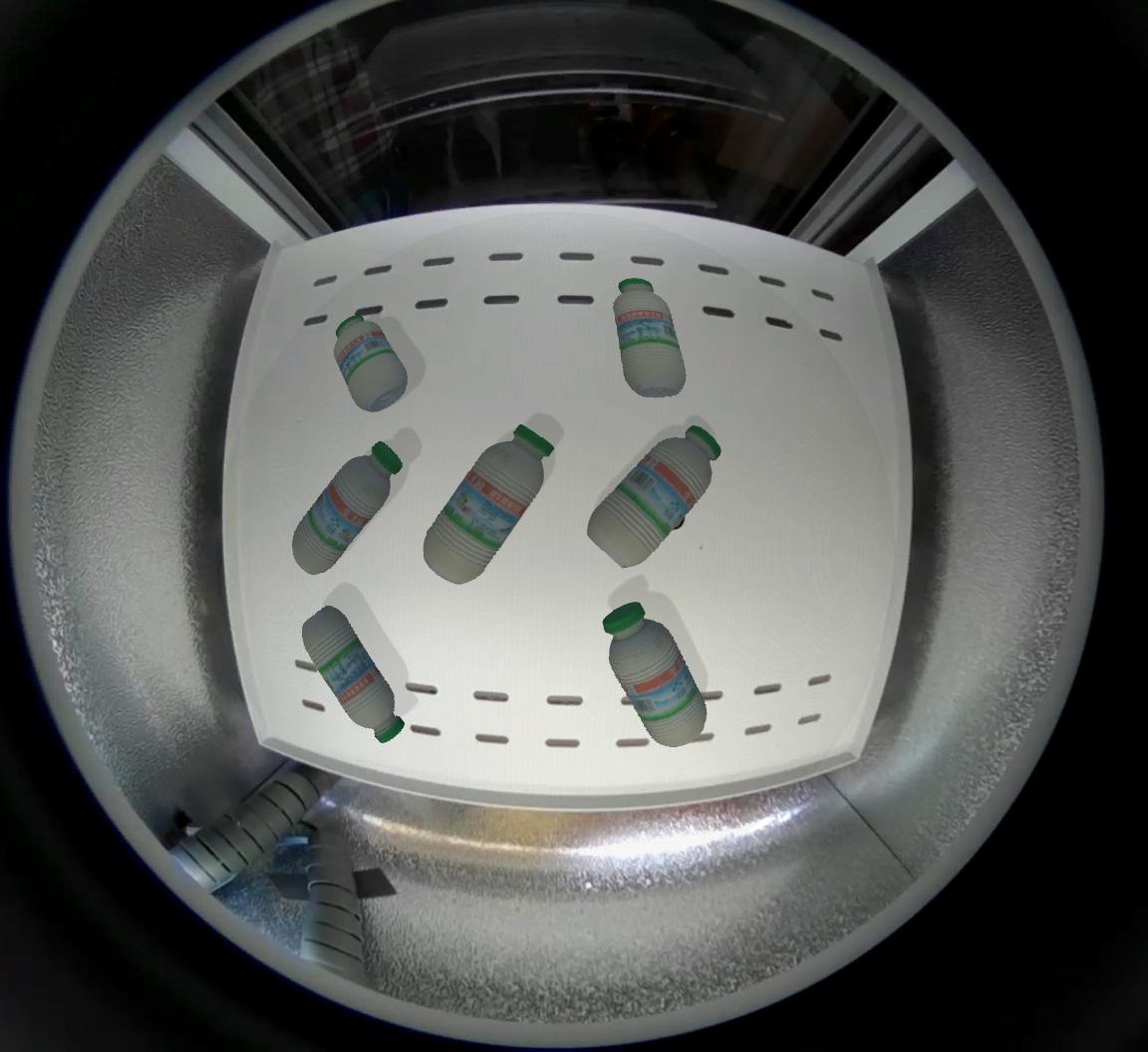}
    \end{minipage}}
  \subfigure[]{
    \centering
    \label{fig:transres:h}
    \begin{minipage}[b]{0.14\textwidth}
      \centering
      \includegraphics[scale=0.08]{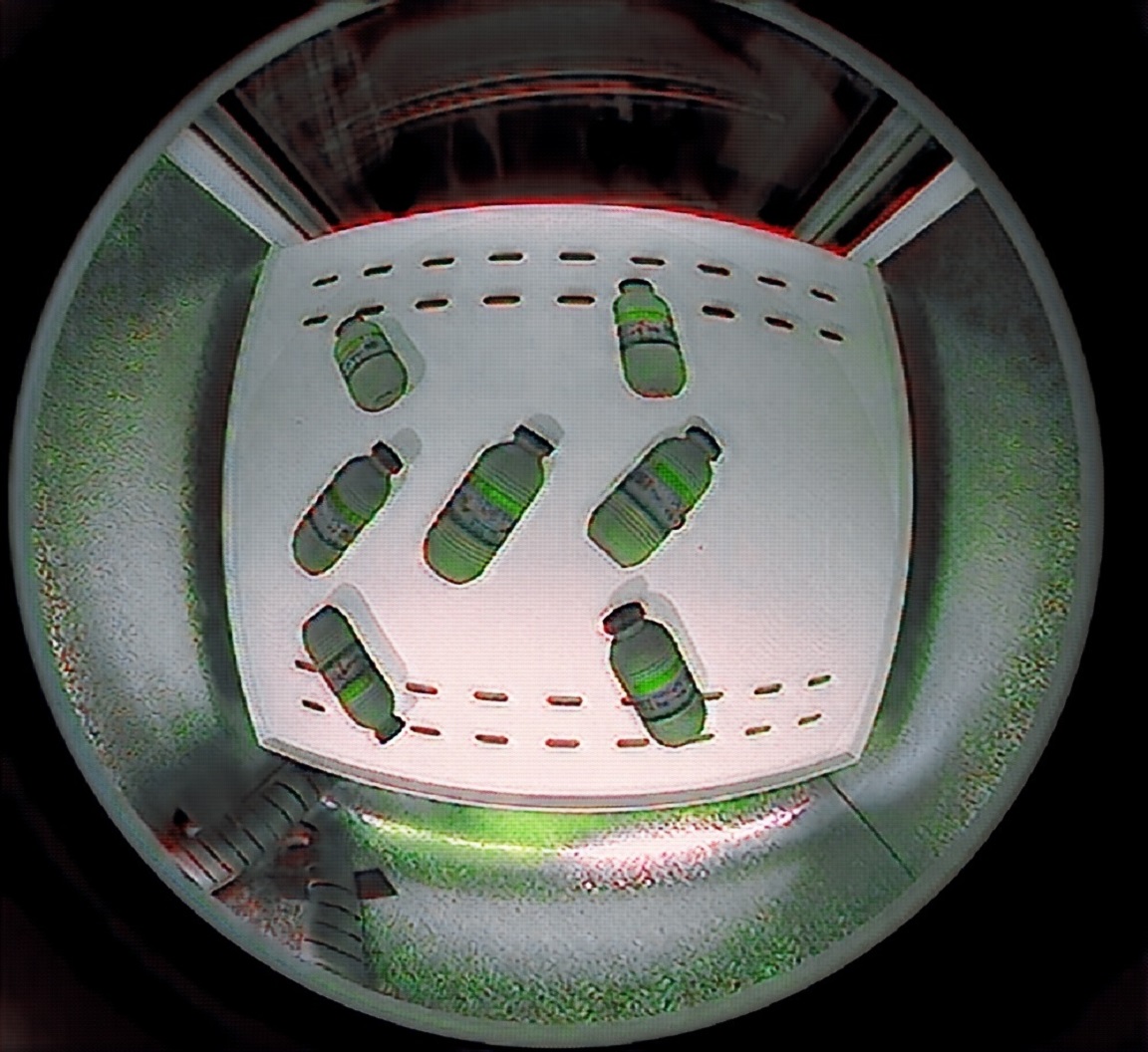}
    \end{minipage}}
  \subfigure[]{
    \centering
    \label{fig:transres:h}
    \begin{minipage}[b]{0.14\textwidth}
      \centering
      \includegraphics[scale=0.08]{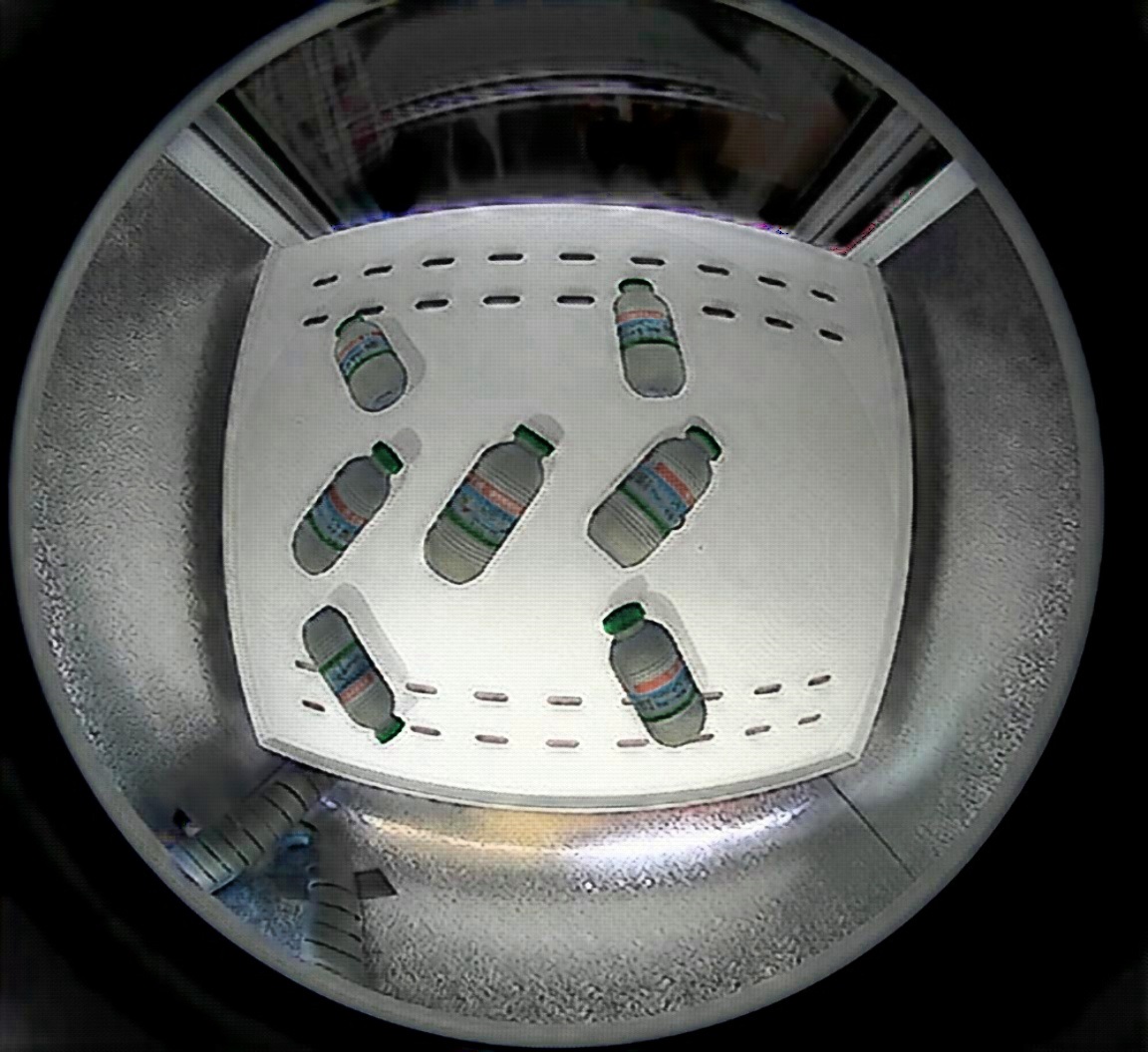}
    \end{minipage}}
  \caption{Results of the virtual-to-real image transfer. (a), (d) and (g): The initially rendered images. (b), (e) and (h): Style-transferred images with the original method. (c), (f) and (i): Images with style transferred in our proposed method.}
  \label{fig:transres}
\end{figure}

The input size original model architecture is $256 \times 256$, however, it¡¯s unable to meet the requirements of accuracy. So the input size is raised to $1000 \times 1000$ in our system. The value of $\lambda_1$, $\lambda_2$ and $\lambda_3$ are set as 10, 3, and 7 respectively in the training step. Figure~\ref{fig:transres} shows some transfer results of original Cycle-GAN and the proposed network.

\subsection{Labeling}

After getting the rendered images, the objects in each image needs to be annotated by computing the bounding rectangle of the visible part of each object and marking its type ID. This may become complicated when the number of objects in the virtual environment increases, as occlusions will usually occur between different objects.

To solve this problem, we adopted a brute force method by setting the textures of the object whose bounds are to be computed to white and the textures of the other objects to black. The background texture is also set to black and all lightings are disabled. The scene will then be rendered and a distorted image with only the visible part of the target object can be obtained. The bounding rectangle can then be calculated with the contour of the white area. As the rendering will become much easier after setting the textures to simple colors and disabling the lights, this computation is quite fast and can be completed in real time.

\section{Experimental evaluation}\label{sec:exp}

\subsection{Experimental environment}

\begin{figure} [htbp]
  \centering
  \subfigure[]{
    \centering
    \label{fig:layout:a} 
    \begin{minipage}[b]{0.22\textwidth}
      \centering
      \includegraphics[scale=0.15]{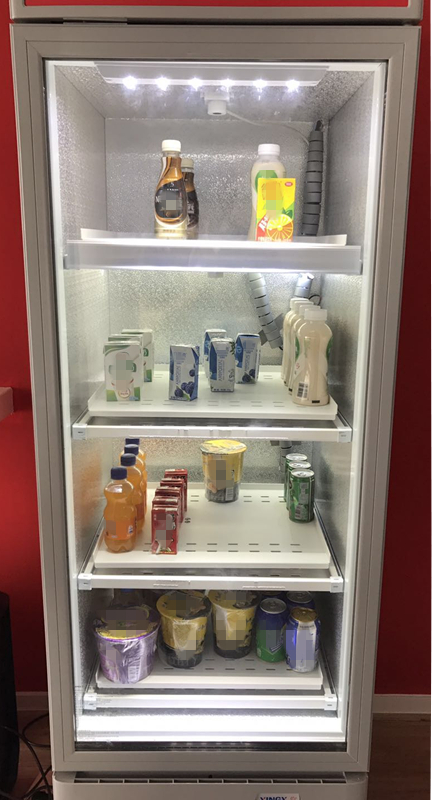}
    \end{minipage}}
  \subfigure[]{
    \centering
    \label{fig:layout:b}
    \begin{minipage}[b]{0.22\textwidth}
      \centering
      \includegraphics[scale=0.6]{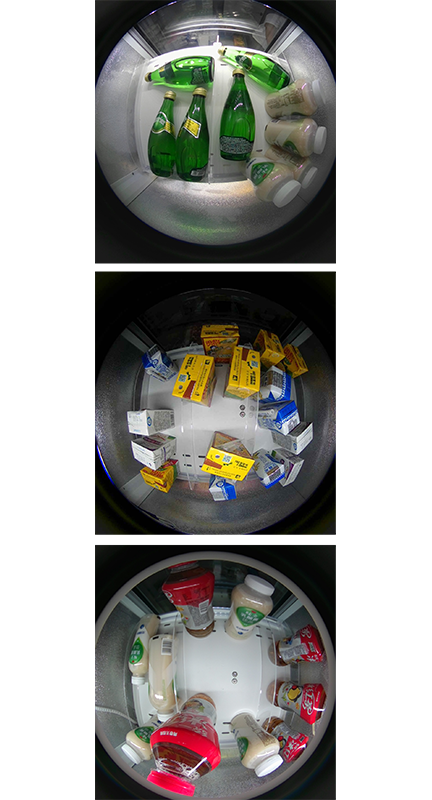}
    \end{minipage}}
  \caption{The smart vending machine used in our experiment. (a)Outside view of the machine. (b)The photos taken from different layers of the machine.}
  \label{fig:hardware}
\end{figure}

We test our approach in a smart vending machine shown in Figure~\ref{fig:hardware}. There are four layers in the machine, each layer is equipped with a fisheye camera for acquiring the commodity images and a fill-in light installed on the upper layer to enhance the stability of environmental lighting. As shown in Figure~\ref{fig:hardware}, the lightning conditions and layer heights vary, and the distortions of object images caused by the fisheye camera are also different.

For object recognition, there are two kinds of schemes, i.e., direct object detection, and general object detection followed by refined classification. In our system, as severe cluster, occlusion and object distortions are introduced by the use of fisheye camera, it is difficult to difficult to make general object detection cover all cases. Therefore, we use direct object detection to recognize commodities inside the vendor. Considering the speed and accuracy factors, PVANET~\cite{KC16}, SSD~\cite{LA16}, and YOLOv3~\cite{RD16} are used for the training and evaluation in this work.

In order to evaluate the performance of our data synthetic method, we trained detectors on both real and synthetic images. For the real image training set, we captured 309 images of 10 types of objects on the first three layers and under different lighting conditions, and labeled 500 bounding boxes for each type of object in those images. For the synthetic training set, we generate totally 400 images, in which 102 different background images, 10 types of illuminations and layer heights are set during the generation process.
There are about 1000 bounding boxes for each reconstructed object in the synthetic images.

The Caffe~\cite{JS14} deep learning framework is used for training PVANET and SSD architectures, and darknet~\cite{RJ13} is used for training of YOLOv3. All the detection results are computed by Intersection over Union (IoU) score on the test set, then mean Average Precision (mAP) in percentage is evaluated.

\subsection{Evaluations}


We first evaluate the performance of the three detection algorithms on training sets with different ratios of the real and synthetic images. We randomly selected 20, 80 and 200 images from the 400 generated images, and there are about 50, 200 and 500 bounding boxes for each type of object in the selected images respectively. These selected images are combined with all the 309 captured images respectively during the training process. For the testing of these trained models, we captured another 308 images of objects on the same three layers and labeled them. There are around 500 bounding boxes for each type of object in the test set. The models trained on the above mentioned datasets are compared with those trained with only the real images on this test set, and the results are shown in Table~\ref{tab:archi}. The 'real' and 'syn' refer to the real and synthetic images, and the numbers in front of them refer to the approximate number of bounding box for each object.

\begin{table}[htbp]
  \caption{Performance evaluation of different architectures.}
  \label{tab:archi}
  \begin{tabular}{ccccc}
    \hline
    \multirow{2}{*}{} & {500 real} & {500 real} & {500 real}& {500 real}\\
     & {} & {+50 syn} & {+200 syn} & {+500 syn}\\
    \hline
    PVANET & 92.72 & 93.11 & 94.43 & \textbf{94.54}\\
    SSD & 88.31 & 88.69 & 89.26 & 90.02\\
    YOLOv3 & 55.31 & 56.67 & 59.42 &62.33\\
  \hline
\end{tabular}
\end{table}

It can be seen that the utilization of the synthetic images during the training improved the performance on the training set remarkably. Among the three algorithms, PVANET achieves the best performance, and the mAP increases nearly 2 percentages with the use of the same number of generated data. This proves that our data generation and adaption method effectively simulated the data distribution in the real environment. It is also worth noting that the precision does not improve much when the number of synthetic images reaches that of the real ones, as the data distribution has been well simulated with that quantity of synthetic data, and more data will not help to enrich that distribution obviously.

\begin{figure} [htbp]
  \centering
  \subfigure[]{
    \centering
    \label{fig:detres:a} 
    \begin{minipage}[b]{0.13\textwidth}
      \centering
      \includegraphics[scale=0.055]{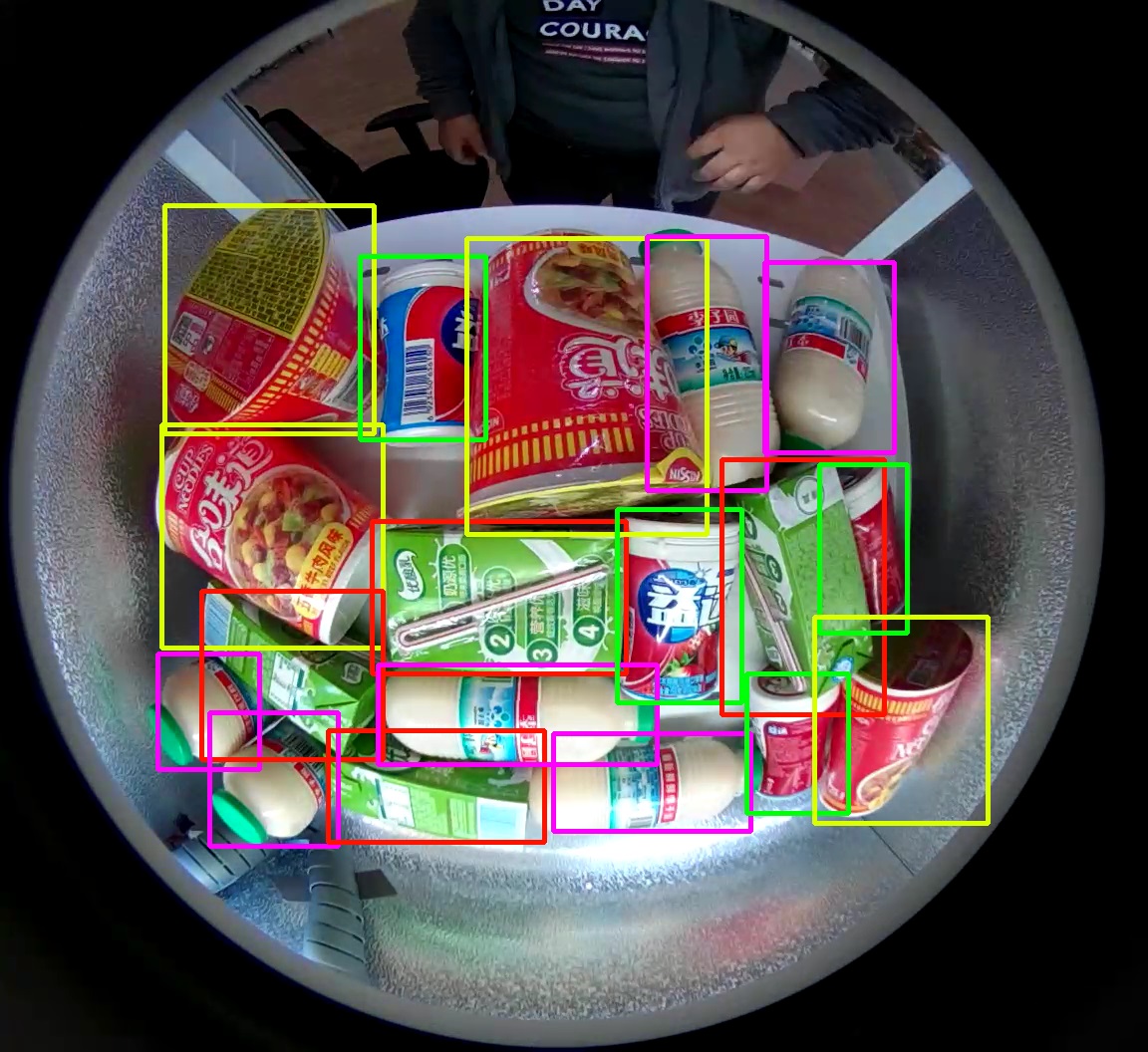}
    \end{minipage}}
  \subfigure[]{
    \centering
    \label{fig:detres:b}
    \begin{minipage}[b]{0.13\textwidth}
      \centering
      \includegraphics[scale=0.055]{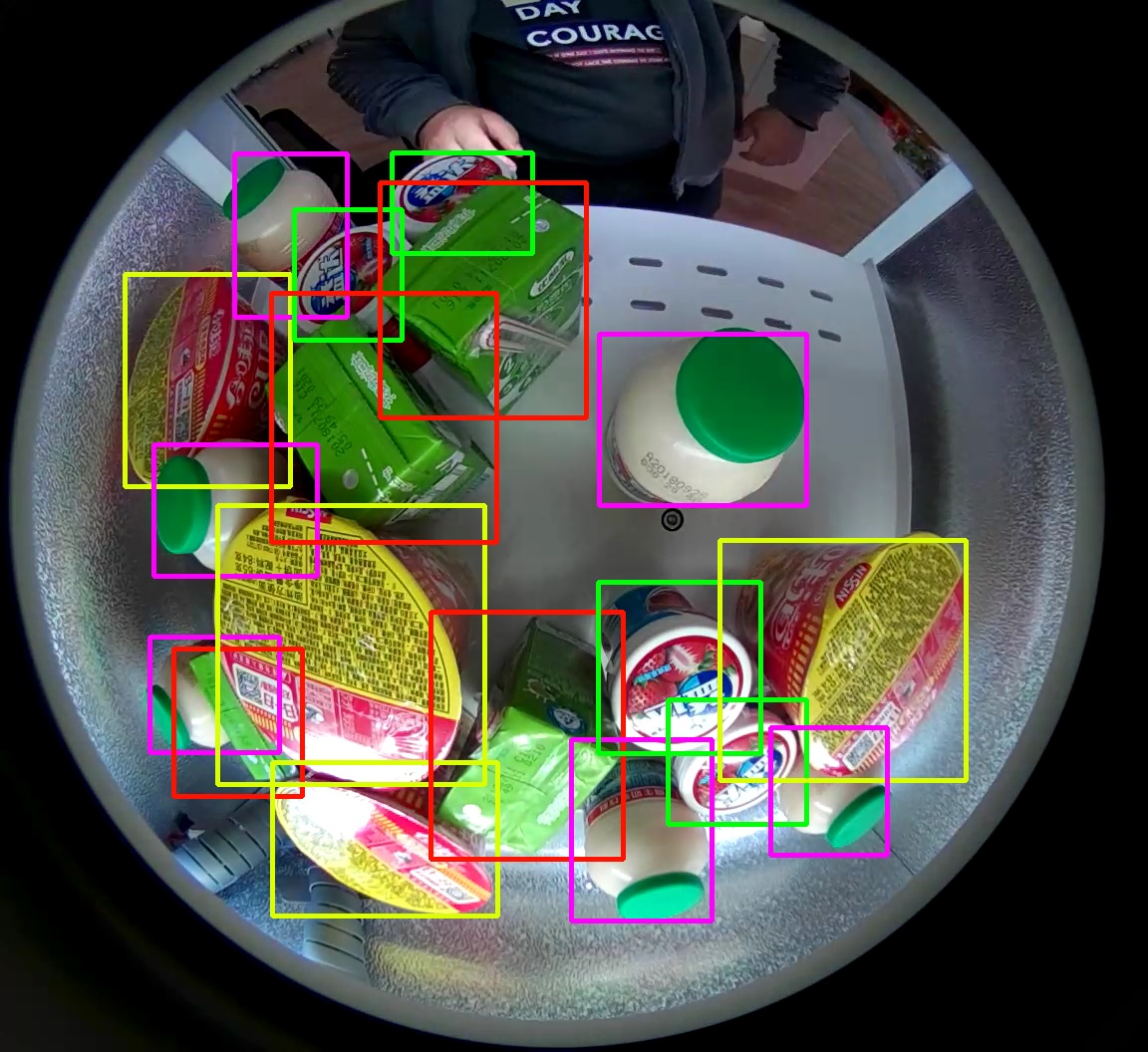}
    \end{minipage}}
  \subfigure[]{
    \centering
    \label{fig:detres:c}
    \begin{minipage}[b]{0.13\textwidth}
      \centering
      \includegraphics[scale=0.074]{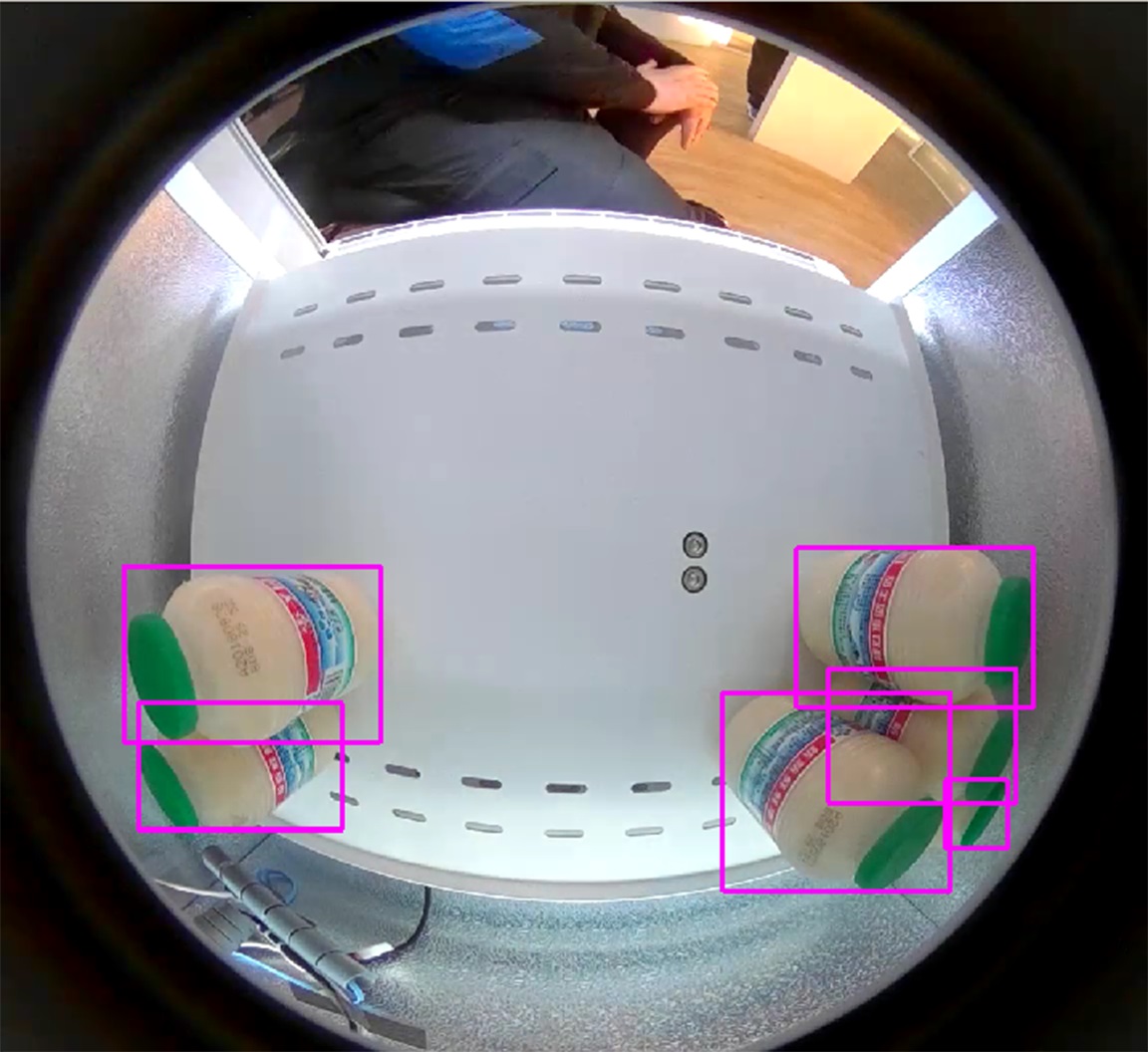}
    \end{minipage}}
  \subfigure[]{
    \centering
    \label{fig:detres:d}
    \begin{minipage}[b]{0.13\textwidth}
      \centering
      \includegraphics[scale=0.074]{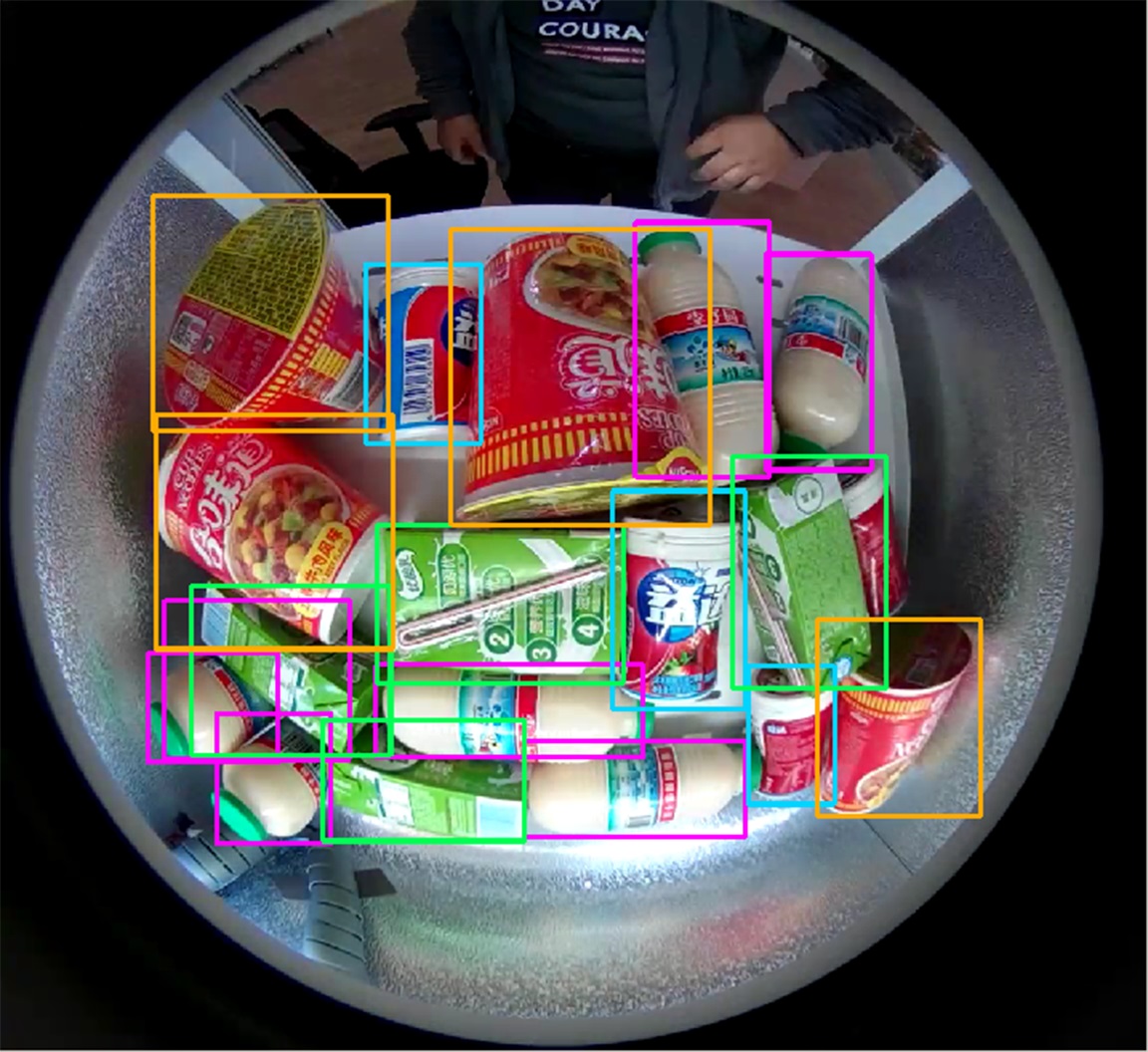}
    \end{minipage}}
  \subfigure[]{
    \centering
    \label{fig:detres:e}
    \begin{minipage}[b]{0.13\textwidth}
      \centering
      \includegraphics[scale=0.074]{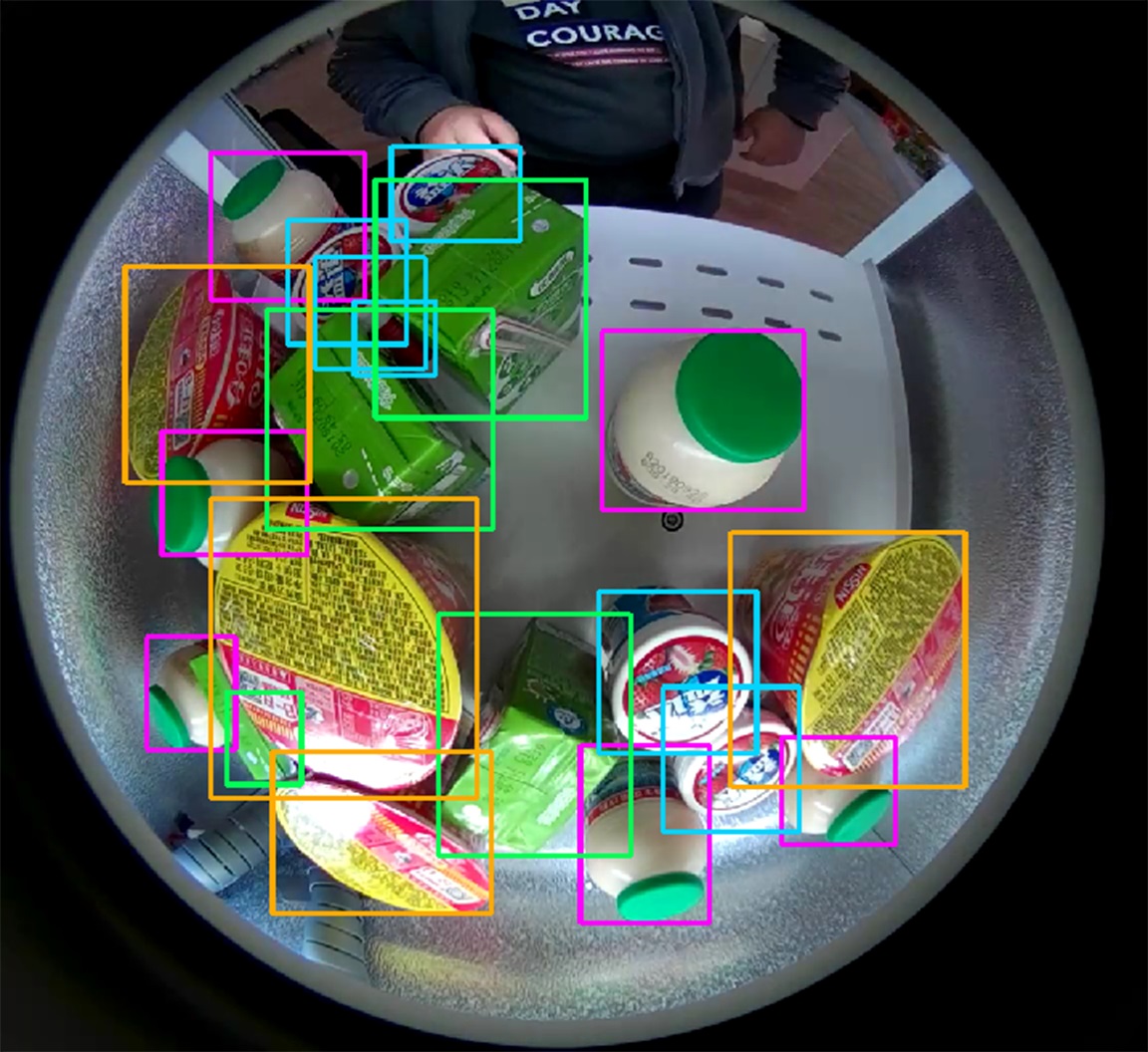}
    \end{minipage}}
  \subfigure[]{
    \centering
    \label{fig:detres:f}
    \begin{minipage}[b]{0.13\textwidth}
      \centering
      \includegraphics[scale=0.055]{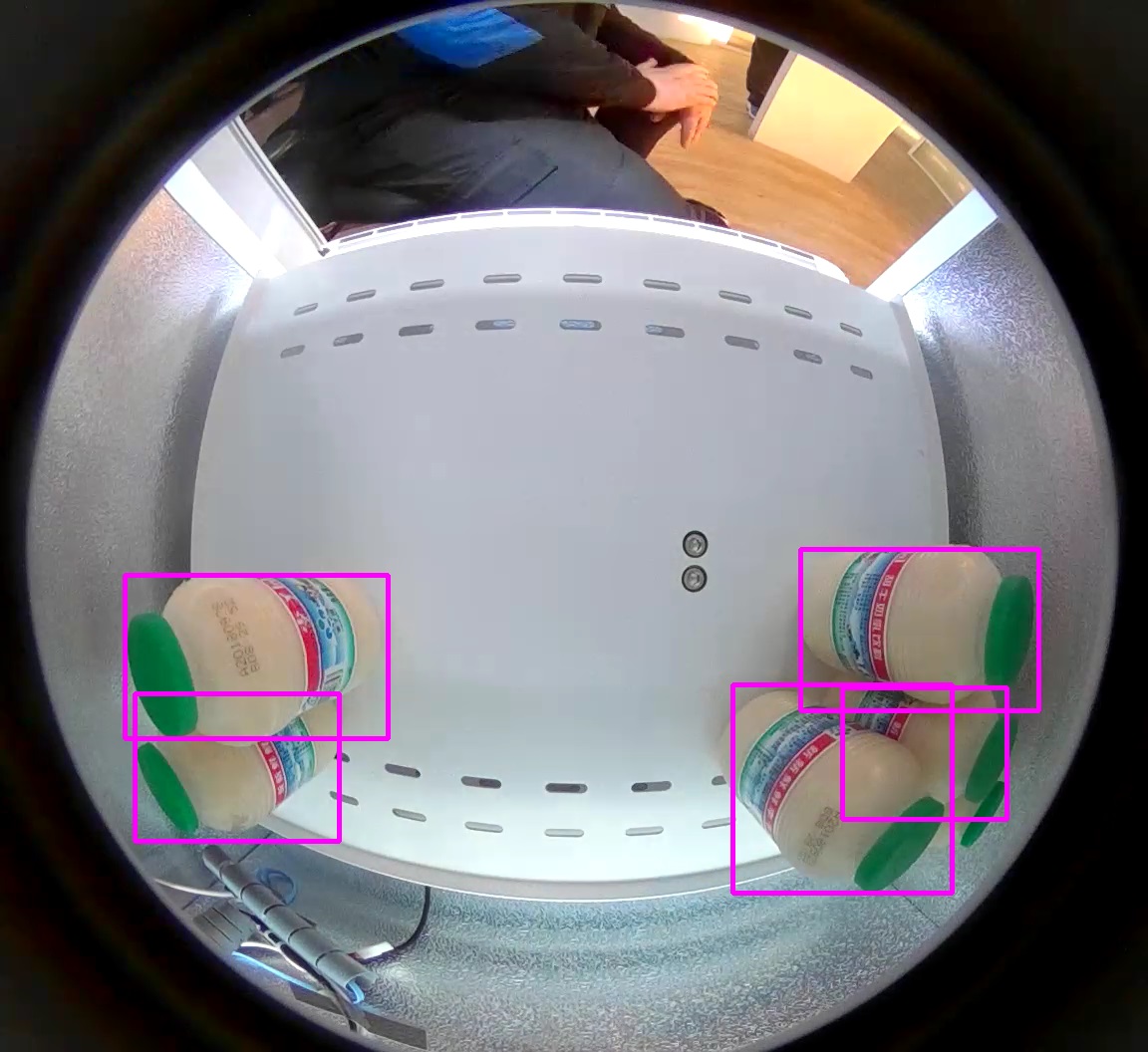}
    \end{minipage}}
  \subfigure[]{
    \centering
    \label{fig:detres:d}
    \begin{minipage}[b]{0.13\textwidth}
      \centering
      \includegraphics[scale=0.074]{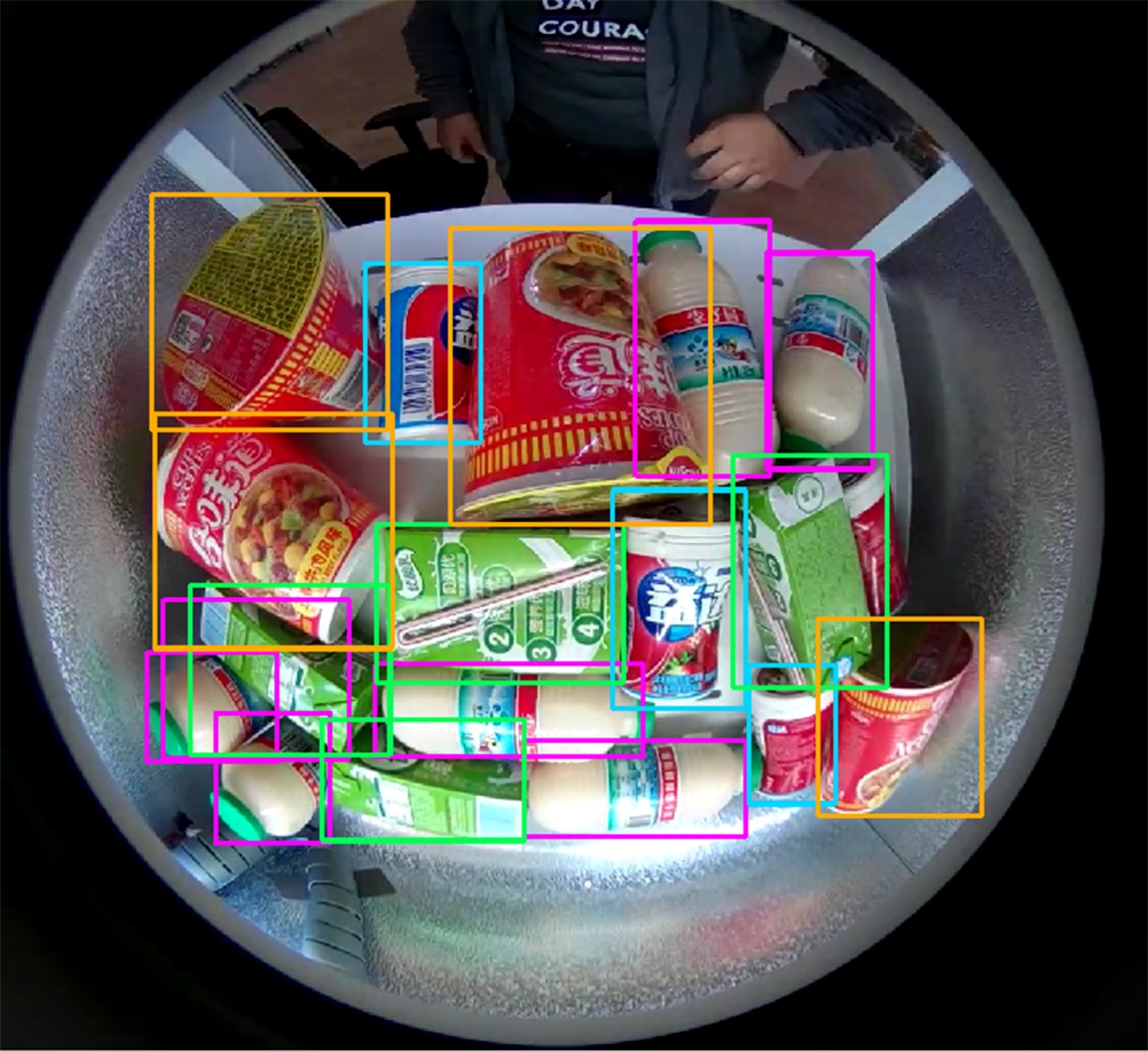}
    \end{minipage}}
  \subfigure[]{
    \centering
    \label{fig:detres:e}
    \begin{minipage}[b]{0.13\textwidth}
      \centering
      \includegraphics[scale=0.074]{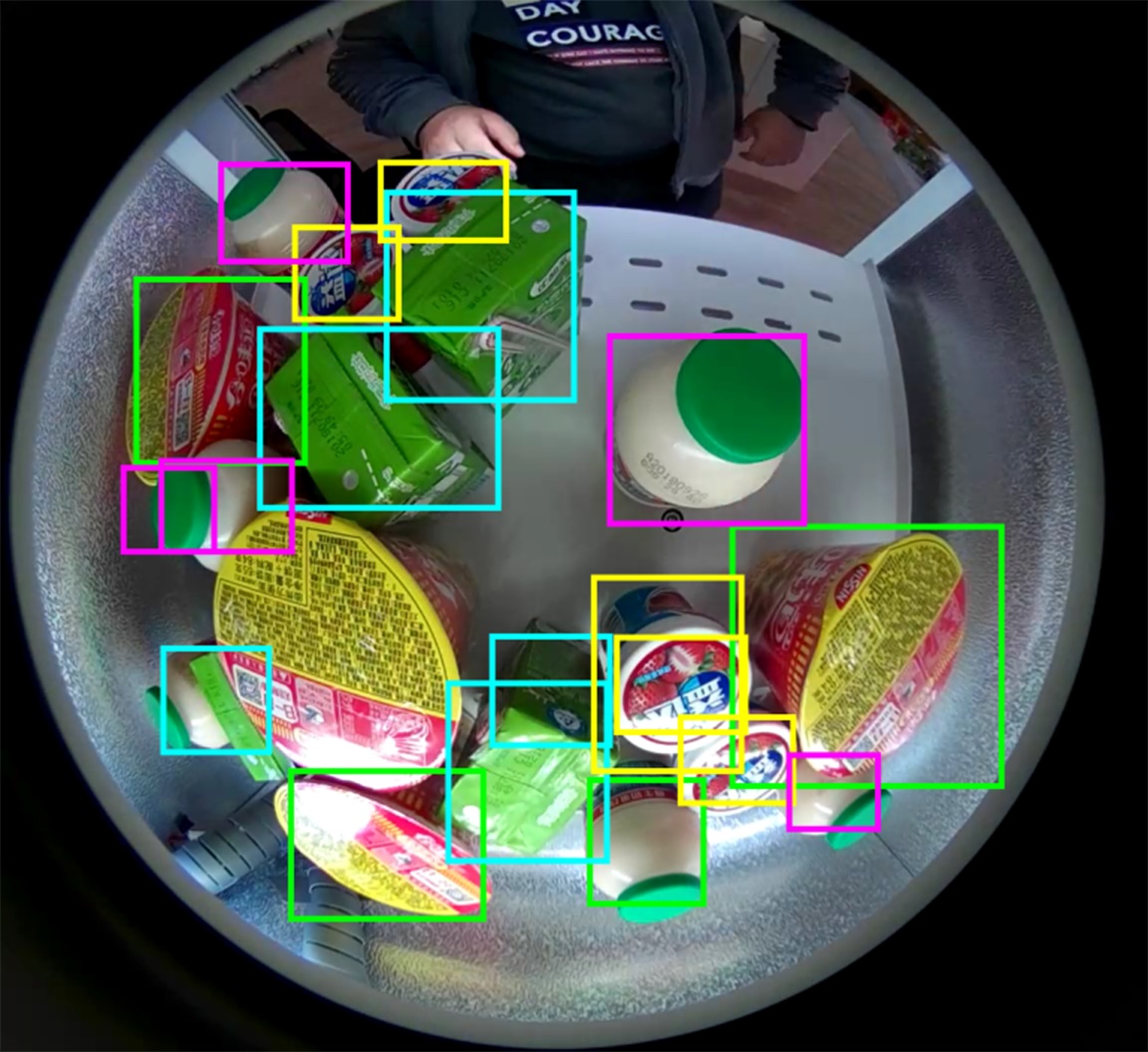}
    \end{minipage}}
  \subfigure[]{
    \centering
    \label{fig:detres:f}
    \begin{minipage}[b]{0.13\textwidth}
      \centering
      \includegraphics[scale=0.074]{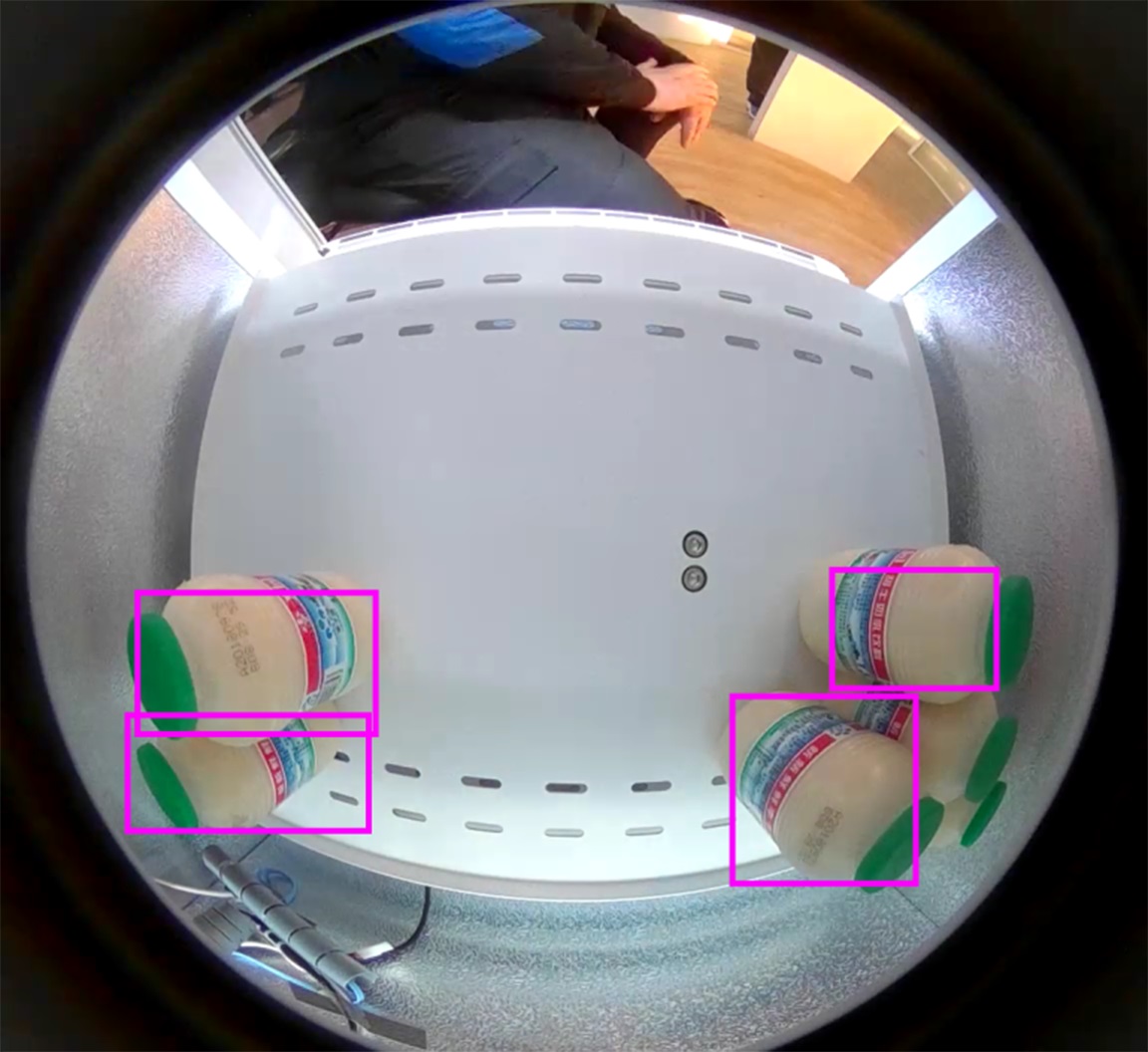}
    \end{minipage}}
  \caption{Detection results of different networks. (a)-(c): Results of using PVANET. (d)-(f): Results of using SSD. (g)-(i): Results of using YOLOv3.}
  \label{fig:detres}
\end{figure}

We show some examples of the detection results of these three architectures trained with the same number of real and synthetic data in Figure~\ref{fig:detres}. It can be seen that SSD and YOLOv3 failed to detect objects that have serious occlusion or shape distortion introduced by fisheye camera.

\begin{figure}[htbp]%
\centering
\includegraphics[scale=0.75]{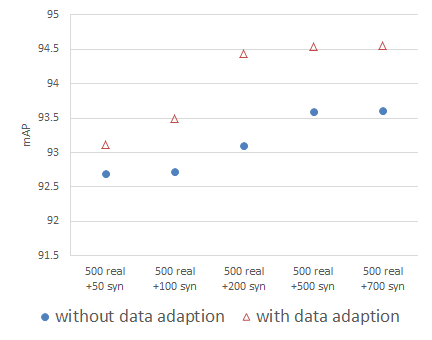}
\caption{The performance comparison of models trained on images generated with and without data adaption.}
\label{fig:adapres}
\end{figure}

To evaluate the effectiveness of the data adaption step in our algorithm, we also trained the PVANET algorithm on datasets containing images generated with and without the style transfer step. Similarly, we combined the real images with 500 bounding boxes per object with synthetic images containing 50, 100, 200, 500, 700 bounding boxes (with and without the style transfer) respectively for the training set. The trained models are also evaluated on the same test set in previous experiment. The mAP results are shown in Figure~\ref{fig:adapres}. It can be seen that the precision of models trained on images with data adaption obviously outperforms those trained on rendered images without further adaption. It is also noted that the result of the data adaption method is able to approach optimal with a quantity of only 200 bounding boxes of each object, while it requires 500 for the generated data without adaption. This proves that our data adaption method can effectively transfer the data distribution of the synthetic images to that of that of the images captured in real environment, and less generated data is thus required, which makes the data generation more effective and less time-consuming.

\begin{figure}[htbp]%
\centering
\includegraphics[scale=0.75]{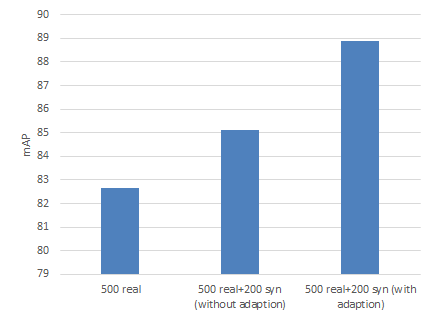}
\caption{Comparison of the generalization capabilities of models trained on real images and those mixed with generated images (with and without adaption).}
\label{fig:genres}
\end{figure}

We also conducted another experiment on a new test set to test the generalization capability of the models trained on the generated images. Specifically, we collected 220 images on the fourth layer of the cabinet, on which the same 10 types of objects are placed and there are also about 500 bounding boxes for each type. PVANET which achieves the best performance is also adopted here. We compared the test results of models trained with the real images of 500 bounding boxes for each object type and the mixed image sets containing 200 synthetic images with and without data adaption. The result is shown in Figure~\ref{fig:genres}. It can be seen that the mAP of models trained on mixed data with initially rendered images has shifted to 85, thanks to the wider range of sampling space covered by the generated images than that of the images taken in the first three layers. The style transfer step further helped to improved the precision to nearly 89. This proves the generalization ability of the models trained with our data generation and adaption method.

From the above evaluation results, it can be seen that by mixing the data generated with our method with a limited number of the real images as the training data, the precision of object detection can be remarkably enhanced in not only the current environment, but also a new environment which has distinct background, lighting and layout conditions. This proves that our data generation and adaption method is able to compensate for the shortage of training data in image object detection tasks, and effectively avoids the resources spent on redundant data generation.

\section{Conclusions}\label{sec:conc}
We presented an automatic data synthesis scheme which is able to produce photo-realistic virtual images of the interior of vending machines for the sake of complementing the training data for object detection within it. Our method is able to simulate the real environment with complex object layout and wide-angle camera with distortions. A novel generative network is further adopted to transfer the intrinsic features of the real images to the synthetic data before the training process such that they will share more common space in terms of data distribution. Experimental results proved that our method is effective on enhancing the object detection precision with the enriched training data and generalizes well into the datasets of new environment.


\bibliographystyle{IEEEtran}
\bibliography{IEEEabrv,egbib}

\begin{thebibliography}{10}
\providecommand{\url}[1]{#1}
\csname url@samestyle\endcsname
\providecommand{\newblock}{\relax}
\providecommand{\bibinfo}[2]{#2}
\providecommand{\BIBentrySTDinterwordspacing}{\spaceskip=0pt\relax}
\providecommand{\BIBentryALTinterwordstretchfactor}{4}
\providecommand{\BIBentryALTinterwordspacing}{\spaceskip=\fontdimen2\font plus
\BIBentryALTinterwordstretchfactor\fontdimen3\font minus
  \fontdimen4\font\relax}
\providecommand{\BIBforeignlanguage}[2]{{%
\expandafter\ifx\csname l@#1\endcsname\relax
\typeout{** WARNING: IEEEtran.bst: No hyphenation pattern has been}%
\typeout{** loaded for the language `#1'. Using the pattern for}%
\typeout{** the default language instead.}%
\else
\language=\csname l@#1\endcsname
\fi
#2}}
\providecommand{\BIBdecl}{\relax}
\BIBdecl

\bibitem{MW16}
C.~Mitash, K.~Wang, K.~E. Bekris, and A.~Boularias, ``Physics-aware
  self-supervised training of cnns for object detection,'' in \emph{IEEE
  International Conference on Robotics and Automation (ICRA)}, 2017.

\bibitem{RB17}
P.~S. Rajpura, R.~S. Hegde, and H.~Bojinov, ``Object detection using deep cnns
  trained on synthetic images,'' \emph{CoRR}, vol. abs/1706.06782, 2017.

\bibitem{TP18}
J.~Tremblay, A.~Prakash, D.~Acuna, M.~Brophy, V.~Jampani, C.~Anil, T.~To,
  E.~Cameracci, S.~Boochoon, and S.~Birchfield, ``Training deep networks with
  synthetic data: Bridging the reality gap by domain randomization,'' in
  \emph{IEEE Conference on Computer Vision and Pattern Recognition Workshops},
  2018, pp. 969--977.

\bibitem{LV15}
A.~Rozantsev, V.~Lepetit, and P.~Fua, ``On rendering synthetic images for
  training an object detector,'' \emph{Computer Vision and Image
  Understanding}, vol. 137, pp. 24--37, 2015.

\bibitem{PJ12}
L.~Pishchulin, A.~Jain, M.~Andriluka, T.~Thorm{\"a}hlen, and B.~Schiele,
  ``Articulated people detection and pose estimation: Reshaping the future,''
  in \emph{IEEE Conference on Computer Vision and Pattern Recognition
  (CVPR)}.\hskip 1em plus 0.5em minus 0.4em\relax IEEE, 2012, pp. 3178--3185.

\bibitem{HB15}
H.~Hattori, V.~N. Boddeti, K.~Kitani, and T.~Kanade, ``Learning scene-specific
  pedestrian detectors without real data,'' in \emph{IEEE Conference on
  Computer Vision and Pattern Recognition (CVPR)}, June 2015, pp. 3819--3827.

\bibitem{HP15}
A.~Handa, V.~Patraucean, V.~Badrinarayanan, S.~Stent, and R.~Cipolla,
  ``Understanding real world indoor scenes with synthetic data,'' in \emph{IEEE
  Conference on Computer Vision and Pattern Recognition (CVPR)}, 2016, pp.
  4077--4085.

\bibitem{RS16}
G.~Ros, L.~Sellart, J.~Materzynska, D.~Vazquez, and A.~M. Lopez, ``The synthia
  dataset: A large collection of synthetic images for semantic segmentation of
  urban scenes,'' in \emph{IEEE Conference on Computer Vision and Pattern
  Recognition (CVPR)}, June 2016, pp. 3234--3243.

\bibitem{TF17}
J.~Tobin, R.~Fong, A.~Ray, J.~Schneider, W.~Zaremba, and P.~Abbeel, ``Domain
  randomization for transferring deep neural networks from simulation to the
  real world,'' in \emph{IEEE/RSJ International Conference on Intelligent
  Robots and Systems (IROS)}, Vancouver, Canada, 2017, pp. 23--30.

\bibitem{MB17}
C.~Mitash, K.~E. Bekris, and A.~Boularias, ``A self-supervised learning system
  for object detection using physics simulation and multi-view pose
  estimation,'' in \emph{IEEE/RSJ International Conference on Intelligent
  Robots and Systems (IROS)}, Vancouver, Canada, 2017, pp. 545--551.

\bibitem{CAR}
A.~Dosovitskiy, G.~Ros, F.~Codevilla, A.~Lopez, and V.~Koltun, ``{CARLA}: {An}
  open urban driving simulator,'' in \emph{The 1st Annual Conference on Robot
  Learning}, 2017, pp. 1--16.

\bibitem{Air}
S.~Shah, D.~Dey, C.~Lovett, and A.~Kapoor, ``Airsim: High-fidelity visual and
  physical simulation for autonomous vehicles,'' in \emph{Field and service
  robotics}.\hskip 1em plus 0.5em minus 0.4em\relax Springer, 2018, pp.
  621--635.

\bibitem{PS15}
X.~Peng, B.~Sun, K.~Ali, and K.~Saenko, ``Learning deep object detectors from
  3d models,'' in \emph{IEEE International Conference on Computer Vision
  (ICCV)}, 2015, pp. 1278--1286.

\bibitem{MK16}
Y.~Movshovitz-Attias, T.~Kanade, and Y.~Sheikh, ``How useful is photo-realistic
  rendering for visual learning?'' in \emph{IEEE European Conference on
  Computer Vision (ECCV) Workshops}.\hskip 1em plus 0.5em minus 0.4em\relax
  Springer International Publishing, 2016, pp. 202--217.

\bibitem{AM17}
H.~A. Alhaija, S.~K. Mustikovela, L.~Mescheder, A.~Geiger, and C.~Rother,
  ``Augmented reality meets deep learning for car instance segmentation in
  urban scenes,'' in \emph{British Machine Vision Conference (BMVC)}, vol.~1,
  2017, p.~2.

\bibitem{HK18}
A.~Hanel, D.~Kreuzpaintner, and U.~Stilla, ``Evaluation of a traffic sign
  detector by synthetic image data for advanced driver assistance systems,''
  \emph{International Archives of the Photogrammetry, Remote Sensing \& Spatial
  Information Sciences}, vol.~42, no.~2, pp. 425--432, 2018.

\bibitem{DC17}
A.~Dai, A.~X. Chang, M.~Savva, M.~Halber, T.~Funkhouser, and M.~Nie{\ss}ner,
  ``Scannet: Richly-annotated 3d reconstructions of indoor scenes,'' in
  \emph{IEEE Conference on Computer Vision and Pattern Recognition (CVPR)},
  2017, pp. 2432--2443.

\bibitem{Mat}
A.~Chang, A.~Dai, T.~Funkhouser, M.~Halber, M.~Nie{\ss}ner, M.~Savva, S.~Song,
  A.~Zeng, and Y.~Zhang, ``Matterport3d: Learning from rgb-d data in indoor
  environments,'' in \emph{International Conference on 3D Vision (3DV)}, 2017,
  pp. 667--676.

\bibitem{AS17}
I.~Armeni, S.~Sax, A.~R. Zamir, and S.~Savarese, ``Joint 2d-3d-semantic data
  for indoor scene understanding,'' \emph{CoRR}, vol. abs/1702.01105, 2017.

\bibitem{GD14}
R.~Girshick, J.~Donahue, T.~Darrell, and J.~Malik, ``Rich feature hierarchies
  for accurate object detection and semantic segmentation,'' in \emph{IEEE
  Conference on Computer Vision and Pattern Recognition (CVPR)}, 2014, pp.
  580--587.

\bibitem{RD16}
J.~Redmon, S.~Divvala, R.~Girshick, and A.~Farhadi, ``You only look once:
  Unified, real-time object detection,'' in \emph{IEEE Conference on Computer
  Vision and Pattern Recognition (CVPR)}, 2016, pp. 779--788.

\bibitem{LA16}
W.~Liu, D.~Anguelov, D.~Erhan, C.~Szegedy, S.~Reed, C.~Y. Fu, and A.~C. Burg,
  ``Ssd: Single shot multibox detector,'' in \emph{IEEE European Conference on
  Computer Vision (ECCV)}, 2016, pp. 21--37.

\bibitem{GR15}
R.~Girshick, ``Fast r-cnn,'' in \emph{IEEE International Conference on Computer
  Vision (ICCV)}, 2015, pp. 1440--1448.

\bibitem{RH17}
S.~Ren, K.~He, R.~Girshick, and J.~Sun, ``Faster r-cnn: Towards real-time
  object detection with region proposal networks,'' \emph{IEEE Transactions on
  Pattern Analysis and Machine Intelligence (TPAMI)}, vol.~39, no.~6, pp.
  1137--1149, 2017.

\bibitem{KC16}
K.-H. Kim, Y.~Cheon, S.~Hong, B.-S. Roh, and M.~Park, ``{PVANET:} deep but
  lightweight neural networks for real-time object detection,'' \emph{CoRR},
  vol. abs/1608.08021, 2016.

\bibitem{Shinning}
``Shining 3d tech.'' \url{http://en.shining3d.com}, accessed Nov 15, 2018.

\bibitem{WH18}
S.~Wu, H.~Huang, T.~Portenier, M.~Sela, D.~Cohen-Or, R.~Kimmel, and M.~Zwicker,
  ``Specular-to-diffuse translation for multi-view reconstruction,'' in
  \emph{IEEE European Conference on Computer Vision (ECCV)}, ser. Lecture Notes
  in Computer Science, vol. 11208, 2018, pp. 193--211.

\bibitem{WZ12}
K.~Wang, J.~Zheng, H.-S. Seah, and Y.~Ma, ``Triangular mesh deformation via
  edge-based graph,'' \emph{Computer-Aided Design and Applications}, vol.~9,
  no.~3, pp. 345--359, 2012.

\bibitem{DN17}
A.~Dai, M.~Nie{\ss}ner, M.~Zollh{\"o}fer, S.~Izadi, and C.~Theobalt,
  ``Bundlefusion: Real-time globally consistent 3d reconstruction using
  on-the-fly surface reintegration,'' \emph{ACM Transactions on Graphics
  (TOG)}, vol.~36, no.~4, p. 76a, 2017.

\bibitem{HD17}
J.~Huang, A.~Dai, L.~J. Guibas, and M.~Nie{\ss}ner, ``3dlite: towards commodity
  3d scanning for content creation.'' \emph{ACM Transactions on Graphics
  (TOG)}, vol.~36, no.~6, pp. 203:1--203:14, 2017.

\bibitem{Unity}
``Unity technologies,'' \url{www.Unity3D.com}, accessed Nov 15, 2018.

\bibitem{FO86}
S.~Fortune, ``A sweepline algorithm for voronoi diagrams,'' in \emph{The Second
  Annual Symposium on Computational Geometry}, 1986, pp. 313--322.

\bibitem{KB06}
J.~Kannala and S.~S. Brandt, ``A generic camera model and calibration method
  for conventional, wide-angle, and fish-eye lenses,'' \emph{IEEE Transactions
  on Pattern Analysis and Machine Intelligence (TPAMI)}, vol.~28, no.~8, pp.
  1335--1340, Aug 2006.

\bibitem{ZP17}
J.~Zhu, T.~Park, P.~Isola, and A.~A. Efros, ``Unpaired image-to-image
  translation using cycle-consistent adversarial networks,'' in \emph{IEEE
  International Conference on Computer Vision (ICCV)}, 2017, pp. 2242--2251.

\bibitem{JS14}
Y.~Jia, E.~Shelhamer, J.~Donahue, S.~Karayev, J.~Long, R.~Girshick,
  S.~Guadarrama, and T.~Darrell, ``Caffe: Convolutional architecture for fast
  feature embedding,'' in \emph{Proceedings of the 22Nd ACM International
  Conference on Multimedia}, 2014, pp. 675--678.

\bibitem{RJ13}
J.~Redmon, ``Darknet: Open source neural networks in c,''
  \url{http://pjreddie.com/darknet/}, 2013--2016.

\end{thebibliography}

\end{document}